\documentclass[lettersize,journal]{IEEEtran}
\usepackage{amsmath,amsfonts}
\usepackage{algpseudocode}
\usepackage{algorithm}
\usepackage{setspace}
\usepackage{array}
\usepackage[font=footnotesize,labelfont=sf,textfont=sf]{caption}
\usepackage{textcomp}
\usepackage{stfloats}
\usepackage{url}
\usepackage{verbatim}
\usepackage{graphicx}
\usepackage{cite}
\usepackage[font=footnotesize,labelfont=sf,textfont=sf]{subcaption}
\usepackage{tabularx}
\usepackage{siunitx}

\graphicspath{{figures/}}

\DeclareRobustCommand{\IEEEauthorrefmark}[1]{\smash{\textsuperscript{\footnotesize #1}}}

\begin{document}

\title{Runtime optimization of acquisition trajectories for X-ray computed tomography with a robotic sample holder}

\author{\IEEEauthorblockN{Erdal Pekel\IEEEauthorrefmark{1,2},
    María Lancho Lavilla\IEEEauthorrefmark{1,2},
    Franz Pfeiffer\IEEEauthorrefmark{2,3,4,5}, and
    Tobias Lasser\IEEEauthorrefmark{1,2}}\\
  \IEEEauthorblockA{\IEEEauthorrefmark{1}Department of Computer Science, School of Computation, Information and Technology, Technical University of Munich, Munich, Germany}\\
  \IEEEauthorblockA{\IEEEauthorrefmark{2}Munich Institute of Biomedical Engineering, Technical University of Munich, Munich, Germany}\\
  \IEEEauthorblockA{\IEEEauthorrefmark{3}Chair of Biomedical Physics, Department of Physics, School of Natural Sciences, Technical University of Munich, Munich, Germany}\\
  \IEEEauthorblockA{\IEEEauthorrefmark{4}Department of Diagnostic and Interventional Radiology, School of Medicine, and Klinikum rechts der Isar, Technical University of Munich, Munich, Germany}\\
  \IEEEauthorblockA{\IEEEauthorrefmark{5}Institute for Advanced Study, Technical University of Munich, Munich, Germany}}

\maketitle

\begin{abstract}
  Tomographic imaging systems are expected to work with a wide range of samples that house complex structures and challenging material compositions, which can influence image quality in a bad way.
  Complex samples increase total measurement duration and may introduce beam-hardening artifacts that lead to poor reconstruction image quality.

  This work presents an online trajectory optimization method for an X-ray computed tomography system with a robotic sample holder.
  The proposed method reduces measurement time and increases reconstruction image quality by generating an optimized spherical trajectory for the given sample without prior knowledge.
  The trajectory is generated successively at runtime based on intermediate sample measurements.

  We present experimental results with the robotic sample holder where two sample measurements using an optimized spherical trajectory achieve improved reconstruction quality compared to a conventional spherical trajectory.
  Our results demonstrate the ability of our system to increase reconstruction image quality and avoid artifacts at runtime when no prior information about the sample is provided.
\end{abstract}

\section{Introduction}\label{sec:introduction-2023}
With the introduction and increasing adoption of robotic manipulators in imaging systems, early adopters expect improved flexibility and image quality.
For X-ray computed tomographic systems in particular, the advantages of a flexible sample holder are obvious: images of the sample can be acquired from different angles more efficiently, ultimately leading to improved reconstruction image quality.
Relevant fields for introducing robotic manipulators for improved imaging results include a wide range of medical applications and non-destructive testing in general.
Robotic manipulators offer the additional benefit of acquiring non-standard trajectories that are not limited in composition and sequence.
In medical scenarios where scan dose reduction benefits the patient, optimized trajectories significantly improve the patient's experience.
In non-destructive testing, where the sample structure and composition are more complex compared to medical use cases, the goal is to improve image quality in the first place.

In recent work, we introduced a flexible robotic arm with seven degrees of freedom (DoF) as a sample holder within a laboratory X-ray computed tomography (CT) setup \cite{Pekel_2022}.
Subsequently, we provided an extensive study on our system's ability to execute spherical trajectories \cite{Pekel_2023}.
The arm adds flexibility to the setup as a sample holder by enabling arbitrary rotations and placement of the sample.
Hence, it allows non-standard trajectories, as opposed to conventional circular or helical trajectories, that are not restricted in their sequence.
We also introduced a suitable calibration mechanism to determine the exact positioning of the sample from the image, as the values reported by the sensors of the robotic arm are not precise enough for reconstruction purposes.
The calibration mechanism requires a sample holder part attached to the robotic arm, which was also introduced in \cite{Pekel_2022}.

In this work, we use the results from our previous publications (see \cite{Pekel_2022,Pekel_2023}) to improve reconstruction image quality further when using the seven DoF robotic arm as a sample holder by introducing optimized spherical trajectories.
We aim to introduce a method that is easy to use and does not require prior knowledge about the provided sample.
In addition, we aim to provide an implementation that runs in real-time without affecting the scanning times achieved when acquiring spherical trajectories with the same robotic sample holder.
In the following, we present our work on trajectory optimization at runtime for unknown samples with the robotic sample holder.
We provide experimental results with two different samples demonstrating the ability of our system to improve reconstruction image quality.
We also provide a quantitative comparison of reconstruction results to conventional spherical trajectories.

In the remainder of this section, we will provide an overview of related work on trajectory optimization methods for imaging systems with and without robotic arms.

Optimized non-circular orbits with no prior knowledge of the 3D geometry of the sample were introduced in \cite{wu2020} for metal artifact avoidance, where the authors presented a method to obtain optimized orbit trajectories for setups with C-arms.
Their method was based on a coarse back projection obtained from low-dose scout views and its subsequent segmentation using a U-Net that was trained on simulated data.
After this segmentation, the X-ray spectral shift for all possible views in the setup was predicted.
The orbit in which this spectral shift was minimized was identified, ensuring the reduction of beam hardening artifacts caused by metal parts.
Our work presents optimized non-circular orbits for a more complex setup with a seven DoF robotic arm rather than a C-arm.
Moreover, the method we present optimizes the robot's trajectory during runtime, using a scout scan as initialization but using all available information after each iteration to improve the optimized trajectory.
Furthermore, our optimization is not based on the X-ray spectral shift, meaning that no information about the absorption coefficient of the material is required.

Another approach for non-circular orbits was presented in \cite{gang2020}.
In this case, the authors used Tuy's condition for data completeness to design orbits that reduced metal artifacts.
An approximate location of the metal in the sample was needed to design these orbits.
The aspect in which our method differs most is the ability to create the trajectory at runtime, leading to an optimized trajectory that is more adjusted to the actual location of the metal parts of the sample.

This work will outline our approach for executing sample-specific trajectories without requiring prior knowledge about the sample.
We will demonstrate the ability of our algorithm to avoid unfavorable acquisition angles when using a robotic arm as a sample holder for arbitrarily complex trajectories.

\section{Methods}\label{sec:methods-2023}
This section discusses the methods for generating optimized trajectories with the robotic arm as a sample holder.
Our system generates the trajectories at runtime without apriori knowledge about the provided sample.
The algorithms implemented herein can be applied to the laboratory environment or in simulation.
After introducing the simulation environment and an overview of the optimization pipeline, we describe more specific aspects like reconstruction volume segmentation, score update, and trajectory pose sampling.
We also refer interested readers to our previous publications \cite{Pekel_2022, Pekel_2023} for a more detailed introduction to the robotic sample holder for X-ray CT.

\subsection{Simulation environment}\label{sec:simulation-environment-2023}
We based our simulation environment on the software package demonstrated in \cite{Pekel_2022} and \cite{Pekel_2023}.
The simulation integrates the FRANKA EMIKA Panda robotic arm with suitable motion planning and X-ray computed tomography components in a unified software package.

Panda is a seven degrees of freedom (DoF) robotic arm with a maximum reach of $855$ mm and a maximum payload of 3 kg.
It has two fingers that can move on a linear axis and grasp objects.
We attached a sample holder to the arm's fingers for calibrating the exact position of the sample on the sinograms with the algorithm developed in \cite{Pekel_2022}.

We modeled the X-ray source and detector from our laboratory setup in the motion planning pipeline to only allow the execution of trajectories where the links of the robotics arm do not interfere with the X-ray beam and hence no occlusions with the sample occur on the detector image.
The realistic modeling of the robotic arm's environment allows us to conclude the applicability of our methods in laboratory conditions.

While the physics of the robotic arm is simulated within the \textit{Gazebo} environment \cite{koenig2004design}, the X-ray measurements are simulated with the tomographic reconstruction framework \textit{elsa} \cite{LasserElsa2019}.
Our simulation environment can execute all steps mentioned in this work in a single run, from trajectory planning to the final reconstruction of the provided sample.
The user can select the sample and sample holder combination that should be simulated from a user interface with a web browser.

\subsection{Pipeline overview}\label{sec:pipeline-overview-2023}
\begin{figure*}[t]
  \centering
  \includegraphics[width=\textwidth]{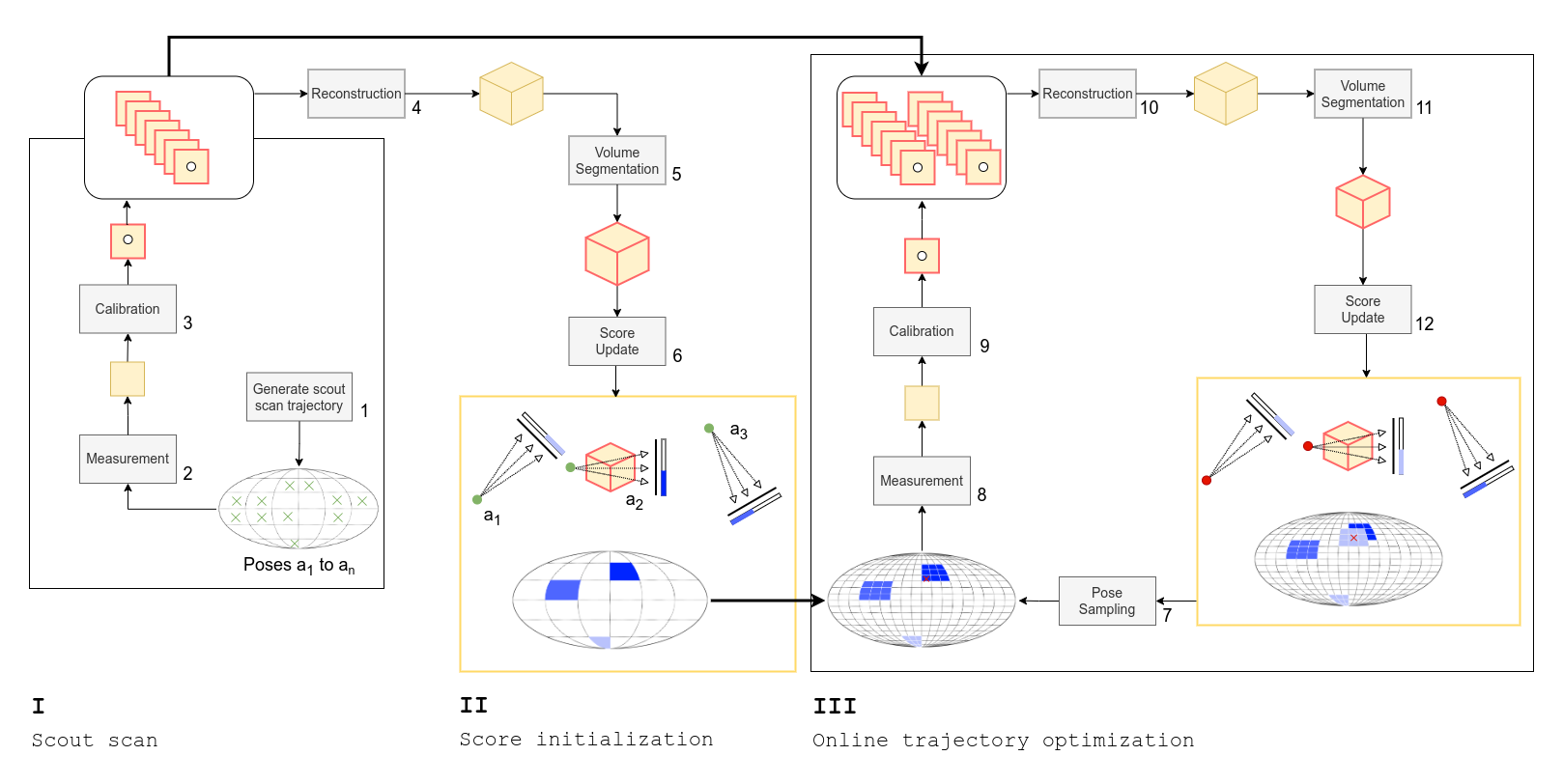}

  \caption{
    \textbf{Online trajectory optimization pipeline}.
    In \textbf{I} (1 - 3), the scout scan procedure starts by generating a spherical trajectory ($a_{1} \dots a_{n}$) that covers the whole sphere with a coarse sampling, only including poses that are reachable by the robotic arm.
    Measurements of the sample are either measured experimentally or simulated with our robotic software package, and the resulting sinograms are calibrated, extracting the exact geometry of the sample on the image.
    In \textbf{II} (4 - 6), the pool of images acquired in \textbf{I} is used to reconstruct the sample in a volume with coarse resolution and to segment highly absorbing regions subsequently.
    Step \textbf{II} is completed by computing a score for each of the poses $a_{i}$ on the coarsely sampled sphere from step \textbf{I}.
    In step \textbf{III} (7 - 12), we enter the open-ended trajectory optimization loop where a densely sampled sphere surface is initialized with the scores from \textbf{II}.
    Likewise, the acquired image pool is initialized with the measurements from \textbf{I}.
    In every iteration of the optimization loop, a new pose is sampled on the sphere surface for measurement, calibration, and volume reconstruction and segmentation by our software package.
    In the final step (12), a score update is initiated on the sphere for the most recently acquired pose and its neighbors.
  }
  \label{fig:oto-pipeline-2023}
\end{figure*}

Figure \ref{fig:oto-pipeline-2023} depicts our trajectory optimization procedure.
The optimization algorithm is split into three parts executed in succession.

In the first step ($\textbf{I}$), the \textit{scout scan} executes a short spherical trajectory and acquires and calibrates images ($2$ and $3$) of the reachable poses.
This trajectory is generated based on a sphere sampling pattern with low density ($1$) to capture the sample from all possible rotations with a minimal set of acquisitions.

In the second step of our algorithm ($\textbf{II}$), the calibrated images from step one are reconstructed ($4$) with a coarse resolution to obtain an approximate representation of the sample.
We segment ($5$) this representation for determining highly absorbing parts of the sample.
Lastly, we calculate a score ($6$) for each successful acquisition angle $a_{1}$ to $a_{n}$.

In the last step of our algorithm ($\textbf{III}$), the pool of successfully acquired images is initialized with the images from the results of the scout scan ($\textbf{I}$).
The sphere data structure is initialized with the scores from the results of the score calculations ($\textbf{II}$).
In contrast to the previous steps ($\textbf{I}$ and $\textbf{II}$), the sphere data structure is initialized with a higher density.
The increased sphere density samples a higher number of poses for more image acquisitions from similar sample angles.
The optimization loop starts by sampling a pose on the sphere and acquiring a sample measurement either experimentally or in simulation based on the current coarse reconstruction.
The resulting image is calibrated and added to the existing pool of images.
The images are used to reconstruct the sample with a coarse resolution ($10$) and segment highly absorbing parts in the next step ($11$), like in the score initialization step (\textbf{II}).
Finally, we recalculate the scores based on the new volume segmentation for all successfully attempted poses up to this point and update the spherical data structure with the new scores ($12$).
The optimization loop continues by sampling a new pose for measurement from the updated spherical map.

The upcoming sections provide detailed descriptions of the individual steps of the online trajectory optimization pipeline.

\subsection{Scout scan}\label{sec:scout-scan-2023}
With the scout scan procedure, we identify the sample's highly absorbing regions on the spherical trajectory before the optimization loop is entered.
By coarsely sampling the sphere surface and generating a short spherical trajectory, we acquire images from all possible angles of the sample in a concise amount of time.
The main advantage of the scout scan procedure is that highly absorbing regions of the sample are detected with a minimal number of acquisitions.
Information about these areas is used in later steps of the optimization pipeline to ensure that the available measurement time is used for less absorbing regions of the sphere.
The acquired images are fed into the trajectory optimization pipeline and serve as an initial set of images for the optimization process.

A vital parameter of the scout scan procedure is the number of poses sampled on the sphere.
The number of poses determines the scout scan's trajectory size and depends on two factors.
First, the sphere discretization number determines the number of equally sized areas into which the sphere surface is cut.
The centers of these areas parametrize the rotations of the target poses, and their sum is the size of the spherical trajectory.
Second, the reachability of the given robotic arm determines which of the sampled poses can be reached by the given robotic arm.
Reachability is a well-defined but complex metric dependent on a range of factors, such as the physical properties of the arm and its operating environment.
We extensively study the reachability of the seven DoF robotic arms for a wide range of operating conditions in our previous work \cite{Pekel_2023}.

We experimented with different values ($12$, $48$, and $108$) and decided to use $108$ for our experiments.
The number cannot be chosen freely as it depends on the grid resolution parameter $N_{side}$ that is required by the HEALPix library, which we use for spherical partitioning.
Given the reachability score of our robotic arm of 82\% \cite{Pekel_2023}, the arm would reach $10$, $40$, and $89$ poses.
Higher numbers could also be considered, but they significantly increase the allocated time for the scout scan procedure; for example, the higher resolution of $N_{side}=4$ results in $192 * 0.82 = 157$ measurements until the optimization loop is entered.

The spherical trajectory of the scout scan is planned by attempting each of the sampled poses individually with the motion planning pipeline.
If the pipeline cannot find a valid and collision-free path to a sampled pose, it is discarded and marked unreachable in the spherical data structure.
Once a valid path is found for the next pose, the robotic arm executes this part of the trajectory.
Once the arm reaches the pose, it stops to acquire an image before attempting to plan a path to the next pose on the sphere.
The acquired images are segmented and calibrated with the methods introduced in \cite{Pekel_2022} to extract the exact geometry of the sample for the reconstruction step.
The calibrated image is added to the pool of successfully acquired images.

\subsection{Score initialization}\label{sec:score-initialization-2023}
In the score initialization step, we perform the operations necessary to transfer the results of the scout scan procedure to the optimization loop.

We reconstruct the sample inside a volume with a coarse resolution with $288 \times 288 \times 288$, which corresponds to a tenth of the detector's resolution.
Choosing a small resolution for the reconstruction allows us to perform the following steps in the pipeline on demand.
The volume segmentation and the score update can be executed in real time when the reconstruction volume has the abovementioned resolution.

The volume segmentation is applied to the resulting reconstruction volume, which contains the sample holder and the sample.
The segmentation output is a volume where the sample holder is filtered out, and only the highly absorbing parts of the sample are visible.
The segmentation is described in detail in section \ref{sec:volume-segmentation-2023}.

The score update step is performed on the output of the volume segmentation for all poses individually.
An absorption score is calculated for each pose, and the resulting score is saved in the spherical data structure for the part of the sphere surface it represents.
The spherical data structure used for this purpose has a low sampling density.
Later, the score will determine the likelihood of the pose's neighborhood to be sampled as the next pose.
The score update step will be discussed in detail in section \ref{sec:score-update-2023}.

In the final step, the scores from the sphere with a low sampling density are transferred to a new sphere with a higher sampling density.
We need the new sphere for sampling more poses from similar angles.
For the score transfer, the score of a specific pose on the sphere with low sampling density is set for all poses it covers on the sphere with higher sampling.

The image pool of the optimization loop is initialized with the measurements from the scout scan procedure. These can provide an initial set of images for the image reconstructions in the first few iterations of the optimization loop.
The reconstructions in the first iterations would have severe artifacts if the image transfer did not occur.

\subsection{Volume segmentation}\label{sec:volume-segmentation-2023}
In order to optimize the acquisition trajectory based on the absorption characteristics of the sample, the projections containing a significant area of the high-absorbing regions of the sample must be avoided.
The first step to achieve this is to identify those areas of the sample in the current reconstruction, which is why volume segmentation is needed.
This segmentation was done by thresholding, aiming to isolate the sample regions whose absorption coefficient is high enough to cause artifacts in the final tomographic reconstruction described in section \ref{sec:reconstruction-2023}.
We chose the threshold in terms of the values of the sample holder and the calibration spheres it contains in the reconstructed volume since these components are strictly necessary for extracting the exact geometry through the calibration process, as presented in \cite{Pekel_2022}, and therefore they are always present.
Moreover, the sample holder is made of a low-absorbing material (polyoxymethylene), so it does not interfere with the reconstruction. At the same time, the spheres are highly absorbing (aluminum) to ensure calibration success.
Both values were extracted from the current reconstruction, using their known location in the reconstructed volume and creating an interval from which the threshold will be picked.
However, to ensure the well-functioning of the algorithm in all scenarios, including if the sample contains parts with higher absorption than the calibration spheres or if the sample does not contain any high-absorbing material, we redefined the upper limit of this interval as the maximum between the spheres value and the maximum value within the sample region in the reconstruction, and the lower limit as the first quartile of the interval.
For the threshold choice, we applied different thresholds within the interval in a decreasing fashion to the reconstruction, the final one chosen as the first one leading to a relevant segmented volume (0.0005 \% of the total volume) or as the lower bound of the interval if it was reached.
The result of the segmentation was a volume of the same size as the reconstruction but only containing values higher than the chosen threshold.
Besides, the sample holder region was blacked out in the segmentation volume to prevent the calibration spheres from being taken as regions to be avoided.

\subsection{Score update}\label{sec:score-update-2023}
In the score update step, we update the surface area parametrized by the given pose with a score.
Ideally, this score should represent the sample's absorption rate when captured from the given pose.
A higher sample attenuation from the given angle should result in a higher score.
We separated the score update into two parts: in the first part, we calculate the score, and in the second part, we find and update the regions on the spherical data structure corresponding to the pose.

The segmented volume is input to the score calculation procedure, which outputs an absorption score for a given pose.
The score is calculated by applying a forward projection to the segmented volume with the calibrated geometry of the given pose and calculating the L0-norm on the resulting sinogram.
The L0 norm effectively calculates the surface area of the non-zero pixels on the detector image.
Since we are applying the forward projection on the segmented volume where solely highly absorbing parts are non-zero, this surface area will be proportional to the amount of high absorption captured from the given pose of the sample.

In the optimization loop (step \textbf{III}), the score is recalculated for all poses that were successfully attempted by our algorithm up to that point.
Recalculating the scores for each pose is necessary as we obtained a new reconstruction volume with the latest successful image acquisition.
The resulting scores are now comparable, and we can create a probability distribution based on these scores in the next step (see section \ref{sec:pose-sampling-2023}).
We are calculating the scores for the poses efficiently by applying a single forward projection on the reconstruction volume with the provided set of angles.
The resulting sinogram contains one slice per pose, and the metric mentioned above (L0-norm) can be calculated on each slice independently.

In the score initialization step, we only set the absorption score of a pose for its surface area on the sphere with low sampling (see Fig. \ref{fig:oto-pipeline-2023} \textbf{II}).

In the trajectory optimization step, we additionally set the scores of all neighbors (see Fig. \ref{fig:oto-pipeline-2023} \textbf{III}, step 12).
We determine the neighbors by searching for all poses on the sphere surface within a predefined distance $r$ radians to the given pose.
This update pattern can be seen on the discretized sphere in Fig. \ref{fig:oto-pipeline-2023} for the red cross in step \textbf{III}, where the immediate spatial neighbors are updated with the light blue color because they fulfill the distance criterion.
Applying the acquired score on the neighbors is necessary as those scores were initialized in the scout scan with a score obtained for a comparably large region.
For example, when dividing the surface area of the sphere into $108$ and $4032$ regions in the scout scan and optimization loop, respectively, the ratio is $41$.
This ratio means a score from the sphere with low sampling density was set for $41$ regions on the sphere with high sampling density in the score initialization step (\textbf{II}).
The distance-based update step ensures that the newly obtained absorption information represented by the new score is spread on the sphere data structure in every iteration.
It is helpful for the pose sampling step, where the score of a pose determines its probability of being picked for the subsequent acquisition.

\subsection{Pose sampling}\label{sec:pose-sampling-2023}
The pose sampling procedure chooses the next pose the robotic arm will approach to acquire a new image.
Ideally, the sampled pose has minimal absorption out of all the remaining poses.

We start with all possible poses on the sphere data structure.
We filter out the ones already attempted in previous runs of the trajectory optimization loop, regardless of successful execution.

Before introducing the probability distribution, we need to determine those poses on the sphere, for which we could not assign a score in the previous steps.
These poses will not be considered in the sampling step as candidates.
Three cases exist when a pose on the sphere results in an undefined state.
The first case is when the pose lies in a region the robotic arm cannot reach.
The second case is when the pose is inside the reachable region of the robotic arm, but the trajectory execution failed due to a hardware error.
The third case is when the pose is reachable, and the trajectory was successfully executed, followed by image acquisition, but the post-processing steps failed.
These steps include segmenting the circles on the image and calibrating the exact geometry of the sample on the image.
Detailed information about the post-processing steps can be found in \cite{Pekel_2022}.

We have illustrated the mapping of a given region on two spheres with low and high sampling in Figures \ref{fig:spherical-trajectories-2023} (\ref{sub@fig:spherical-mapping-low-density-marker-spot-2023}) and (\ref{sub@fig:spherical-mapping-high-density-marker-spot-2023}).
It is worth noting that a pose $p_{i}$ that was not reachable for the arm in the scout scan procedure is not sampled in the probability distribution defined in the next section of our implementation.
In contrast, if a pose $p_{i}$ was reachable, we initialize the poses $p_{i1}$ to $p_{i9}$ inside $p_{i}$ with the score computed for $p_{i}$ in the score initialization step (\textbf{II}) as illustrated in Fig. \ref{fig:spherical-trajectories-2023}(\ref{sub@fig:spherical-mapping-high-density-marker-spot-2023}).
This is a reasonable strategy for cases where measurement time is valuable.
Another valid strategy is to initialize $p_{i1}$ to $p_{i9}$ with a score greater than zero, assigning sampling probabilities greater than zero for these poses in the next step.
This is a more time-consuming strategy, but it is also valid as the robotic arm might still be able to reach the borders of the unreachable area represented by $p_{i}$.

In the next step, we create a discrete probability distribution where the scores for each pose are weighted with an inverse weighting scheme:
Poses with a higher absorption score correspond to a lower weight in the probability distribution and vice versa.
Applying an inverted weighting scheme ensures that poses with high absorption scores are less likely to be drawn from the probability distribution.
Moreover, poses without a score are assigned weight $0$, meaning that they will not be drawn since their probability equals $0$.
This happens when the neighborhood of the pose could not be reached by the robotic arm in the scout scan, when the sphere was sampled with lower density, summing multiple poses into one pose.

We use the parameter $s$ on the inverted weights to model a penalty on the absorption scores:
\begin{equation}\label{eq:score-weight-2023}
  w_{i} = \frac{1}{x^{s}},
\end{equation}

where $s \geq 1$ is the penalty parameters and $x$ is the absorption score.
The score is raised to the power of $s$ to place a sufficiently high penalty on highly absorbing poses.
Different choices for this parameter ($s=\{5,10,30\}$) are displayed in Figures \ref{fig:hyperparameters-trajectories-2023} (\subref{fig:hyperparameters-r5-s5-2023} to \subref{fig:hyperparameters-r15-s20-2023}) for experiments with sample number 1.

The resulting weights are used to create a discrete probability distribution with the C++ standard library component \textit{std::discrete\_distribution}.
The probability of each weight element $w_{i}$ is calculated with $p(i) = \frac{w_{i}}{S}$, where $S = \sum_{i} w_{i}$ is the sum of all weights.
We sample the next valid pose from the discrete distribution with the high-entropy random number engine \textit{std::mt19937} initialized with an entropy-based seed provided by \textit{std::random\_device}.

\subsection{Reconstruction}\label{sec:reconstruction-2023}
We used around $725$ equidistant X-ray projections for the tomographic reconstruction along three different spherical trajectories sized $720 \times 720$ pixels with a spacing of $600 \: \mu m$.
For the different trajectories, the number of projections varied by $+4.8$\%: the whole sphere trajectory successfully acquired and calibrated 725 images, the random trajectory 748, and the optimized trajectory 760.
The reconstruction volume consisted of $720 \times 720\times 720$ isotropic voxels with a $38 \: \mu m$ spacing.
We used our C++ reconstruction framework \textit{elsa} \cite{LasserElsa2019} to perform the reconstruction using an iterative conjugate gradient solver run for $30$ iterations on a Tikhonov regularized weighted least squares problem, with the Josephs method for X-ray transform discretization and cone beam geometry.
Further iterations showed no improvement in the cost function.

\subsection{Software stack}\label{sec:software_stack-2023}
The central part of our software stack is the \textit{Robot Operating System} (\textit{ROS}) \cite{quigley2009ros}, which is a middleware for the communication of independent processes across a network.
We accomplish robot manipulation with the \textit{MoveIt!} framework \cite{coleman2014reducing,moveit-web} and the \textit{franka\_ros} configuration package \cite{frankaros}.
For image processing tasks we use \textit{OpenCV} \cite{opencv_library}, for multithreading on the CPU \textit{OpenMP} \cite{Dagum1998OpenMPAI} and for the tomographic reconstruction \textit{elsa} \cite{LasserElsa2019}.
The sphere discretization in section \ref{sec:methods-2023} was implemented with the HEALPix C++ and Python interfaces \cite{Gorski2005, Zonca2019}.
The 2D spherical projection maps were generated using matplotlib, and basemap \cite{Hunter:2007}.

\section{Experiments and results}\label{sec:experiments-and-results-2023}
We designed different experiments to evaluate the performance of our trajectory optimization algorithm.
The first set of experiments analyzes the sensitivity of our algorithm to the hyperparameters introduced in sections \ref{sec:score-update-2023} and \ref{sec:pose-sampling-2023}.
The second set of experiments comprehensively assesses the reconstruction image quality for two samples.
We also provide a quantitative analysis of the reconstruction images compared to conventional spherical trajectories.
The experiments in this section were executed in our simulation environment (see section \ref{sec:simulation-environment-2023}).

\subsection{Trajectory optimization parameters}\label{sec:experiments-score-update-parameters-2023}
We have conducted experiments for the two hyperparameters of our algorithm mentioned in section \ref{sec:methods-2023}.
The disk radius parameter $r$ mentioned in section \ref{sec:score-update-2023} determines the size of the neighborhood that is affected by the new score of the given pose.
A higher choice for $r$ will apply a more significant update on the sphere surface.
The score penalty parameter $s$ mentioned in section \ref{sec:pose-sampling-2023} penalizes the absorption rate of the given pose.
A higher choice for $s$ will result in a lower weight of the pose in the probability distribution and hence a lower probability of being sampled for image acquisition.
We used three values for each hyperparameter: $r=\{5,10,15\}$ and $s=\{5,10,20\}$.
Our goal with these experiments was to analyze the sensitivity of our algorithm to different choices for the hyperparameters.

\begin{figure*}[t]
  \centering
  \subfloat[$r=5$, $s=5$]{\includegraphics[width=.3\textwidth]{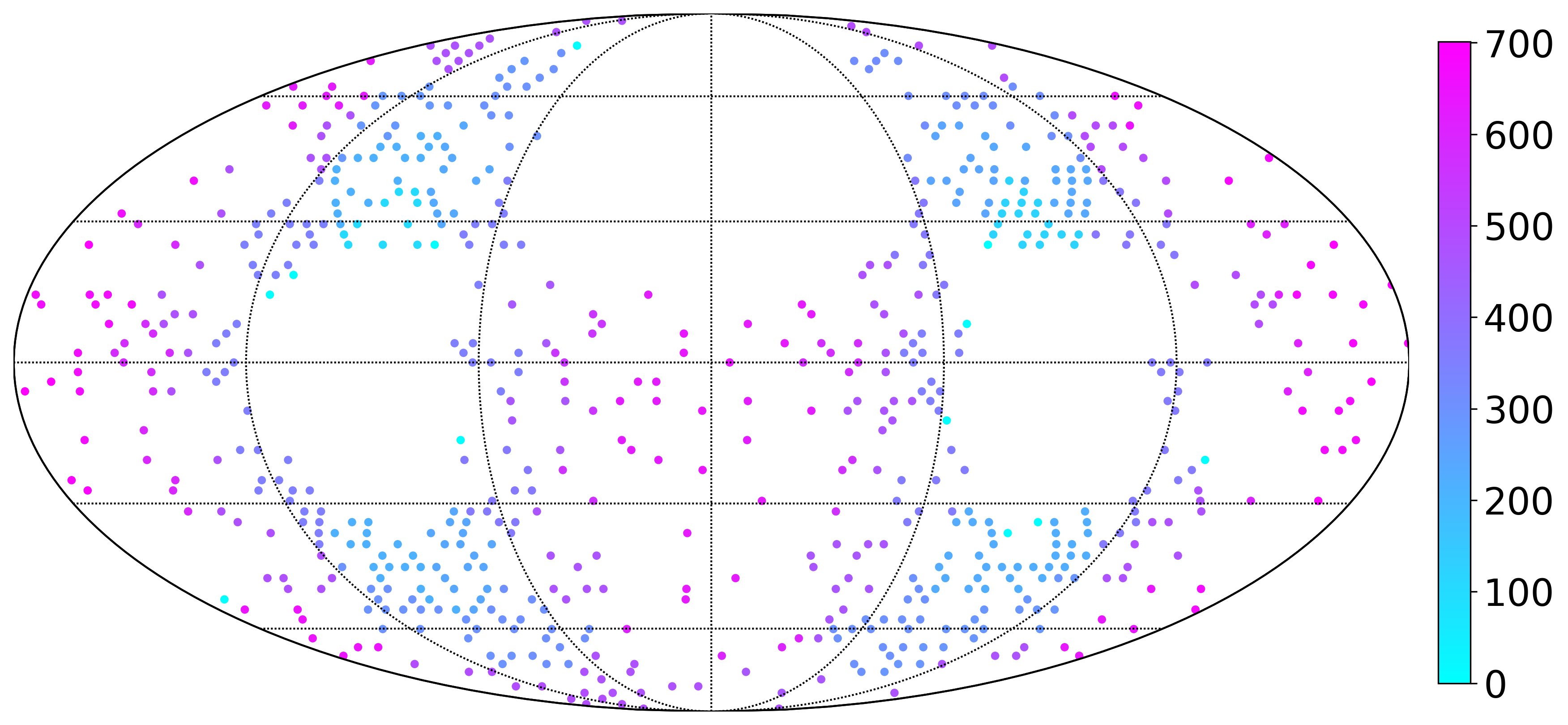}
    \label{fig:hyperparameters-r5-s5-2023}
  }
  \hfil
  \subfloat[$r=5$, $s=10$]{\includegraphics[width=.3\textwidth]{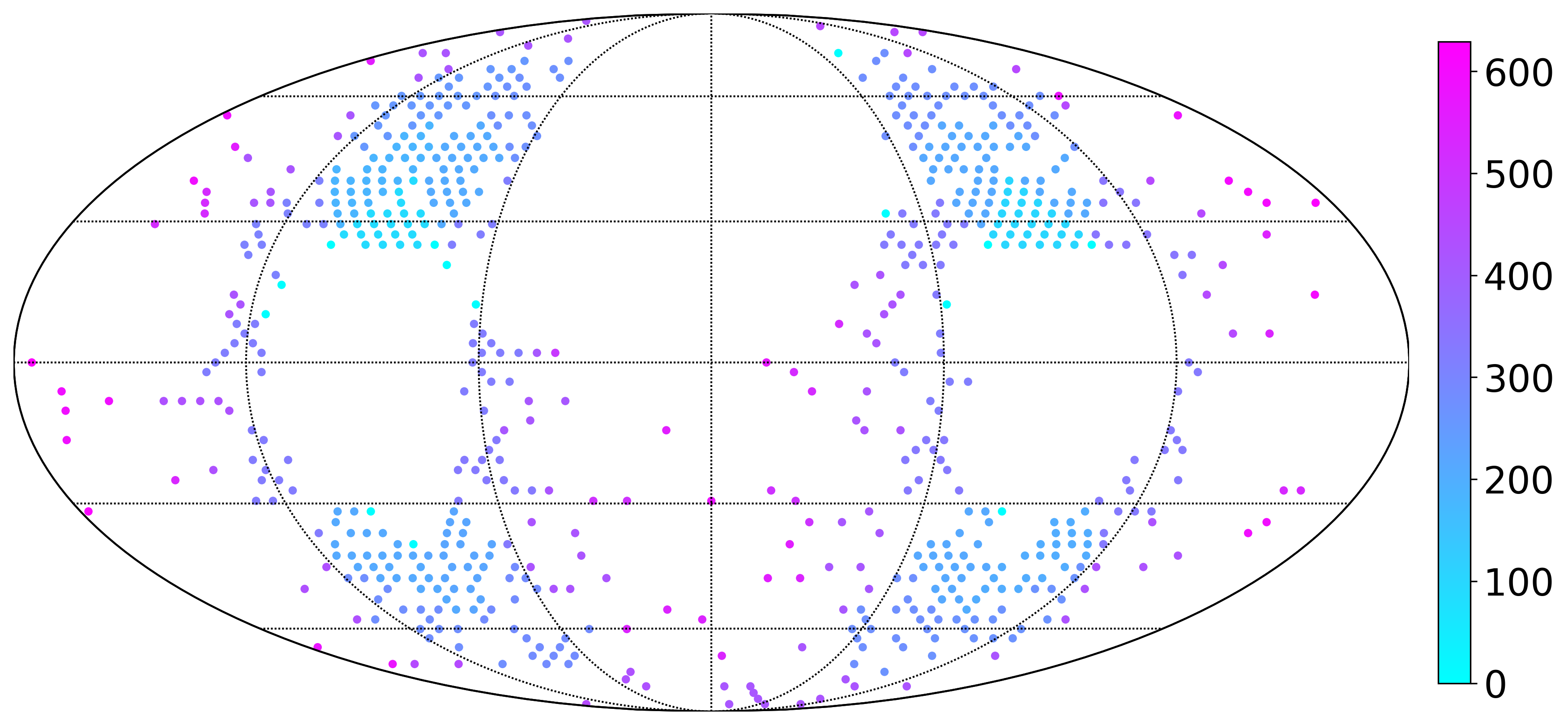}
    \label{fig:hyperparameters-r5-s10-2023}
  }
  \hfil
  \subfloat[$r=5$, $s=20$]{\includegraphics[width=.3\textwidth]{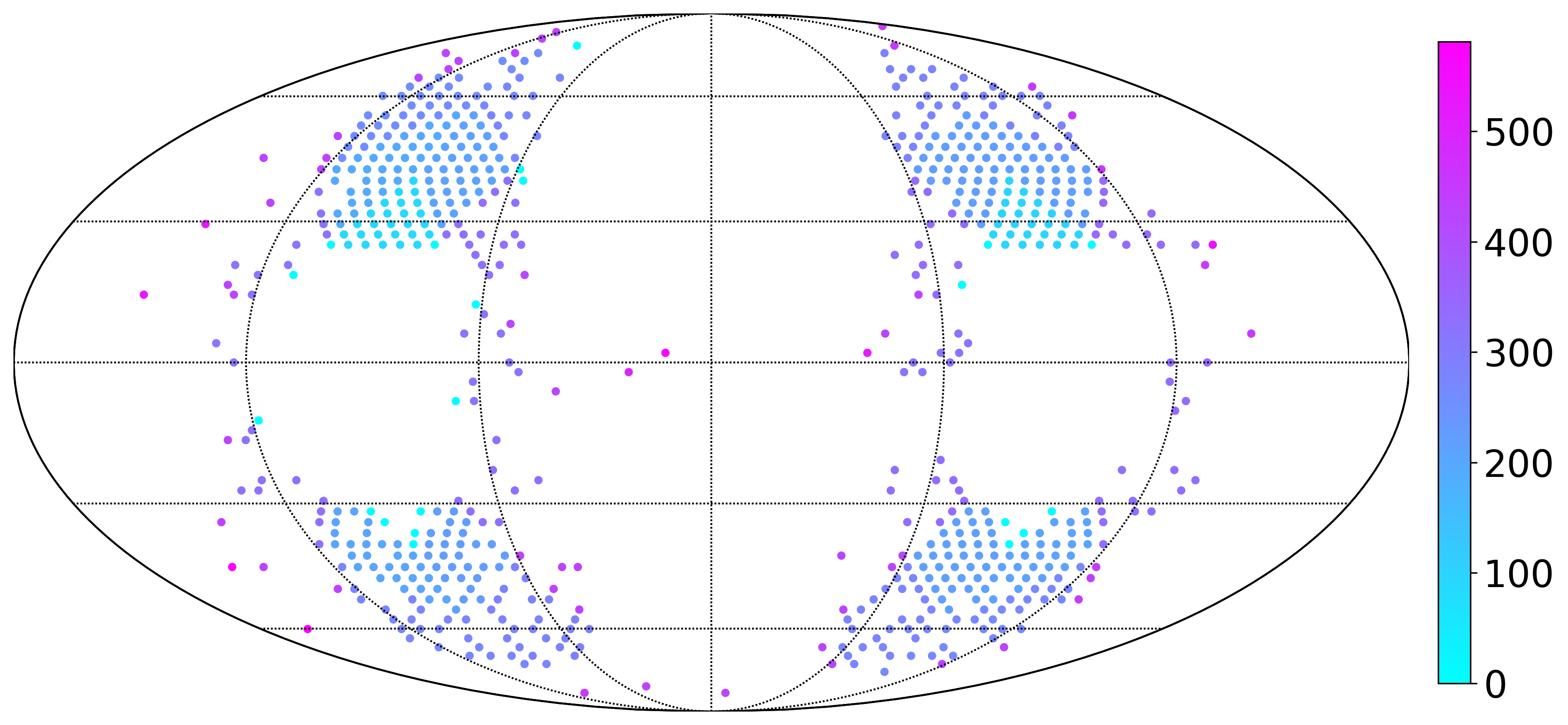}
    \label{fig:hyperparameters-r5-s20-2023}
  }

  \subfloat[$r=10$, $s=5$]{\includegraphics[width=.3\textwidth]{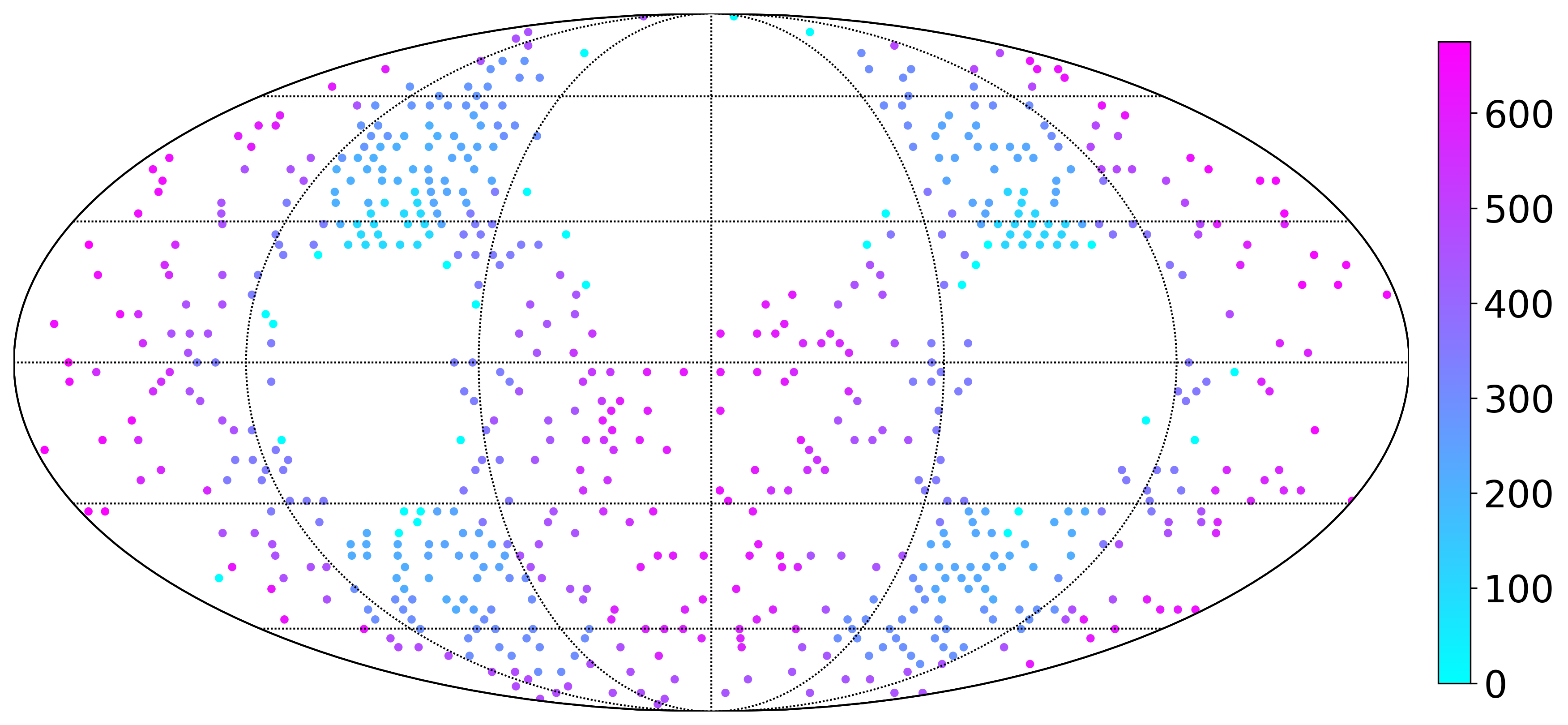}
    \label{fig:hyperparameters-r10-s5-2023}
  }
  \hfil
  \subfloat[$r=10$, $s=10$]{\includegraphics[width=.3\textwidth]{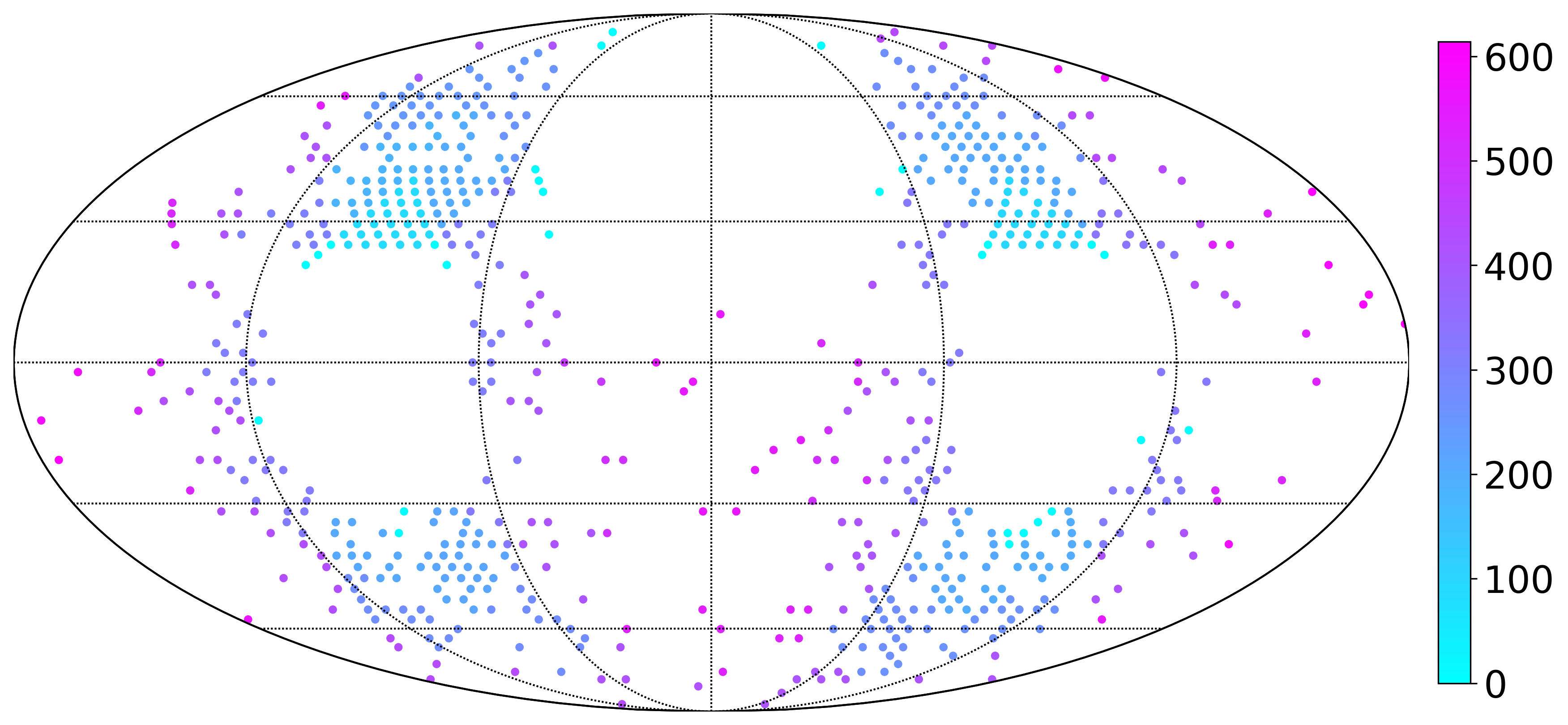}
    \label{fig:hyperparameters-r10-s10-2023}
  }
  \hfil
  \subfloat[$r=10$, $s=20$]{\includegraphics[width=.3\textwidth]{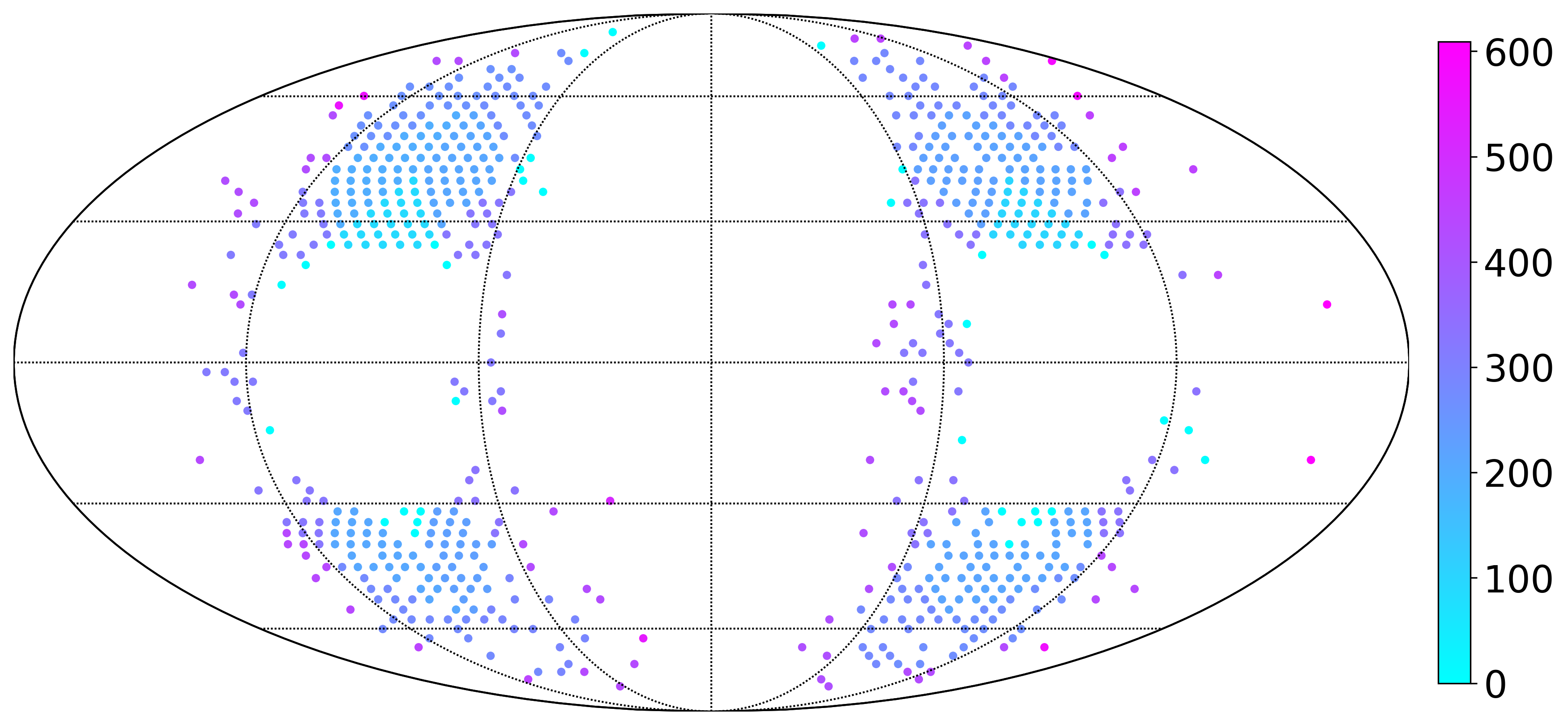}
    \label{fig:hyperparameters-r10-s20-2023}
  }

  \subfloat[$r=15$, $s=5$]{\includegraphics[width=.3\textwidth]{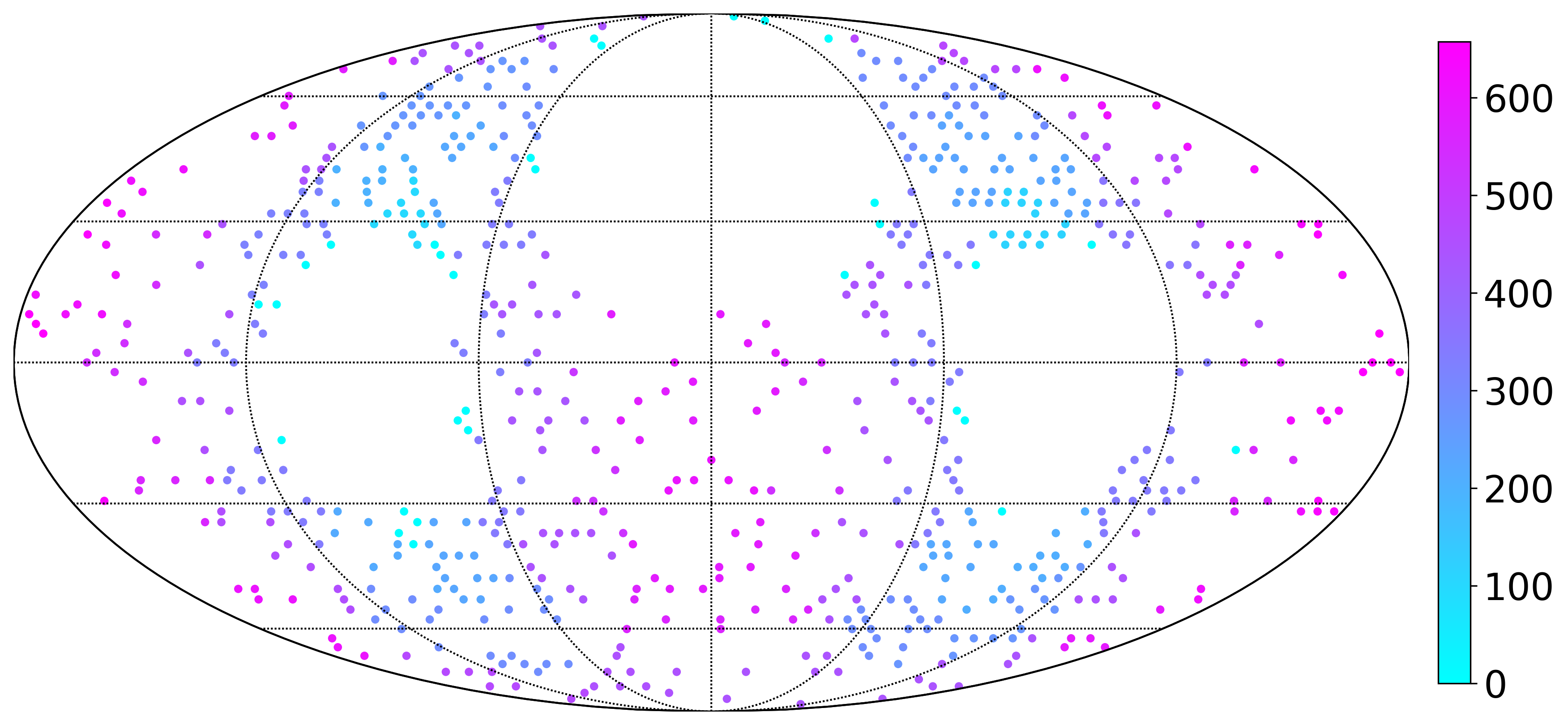}
    \label{fig:hyperparameters-r15-s5-2023}
  }
  \hfil
  \subfloat[$r=15$, $s=10$]{\includegraphics[width=.3\textwidth]{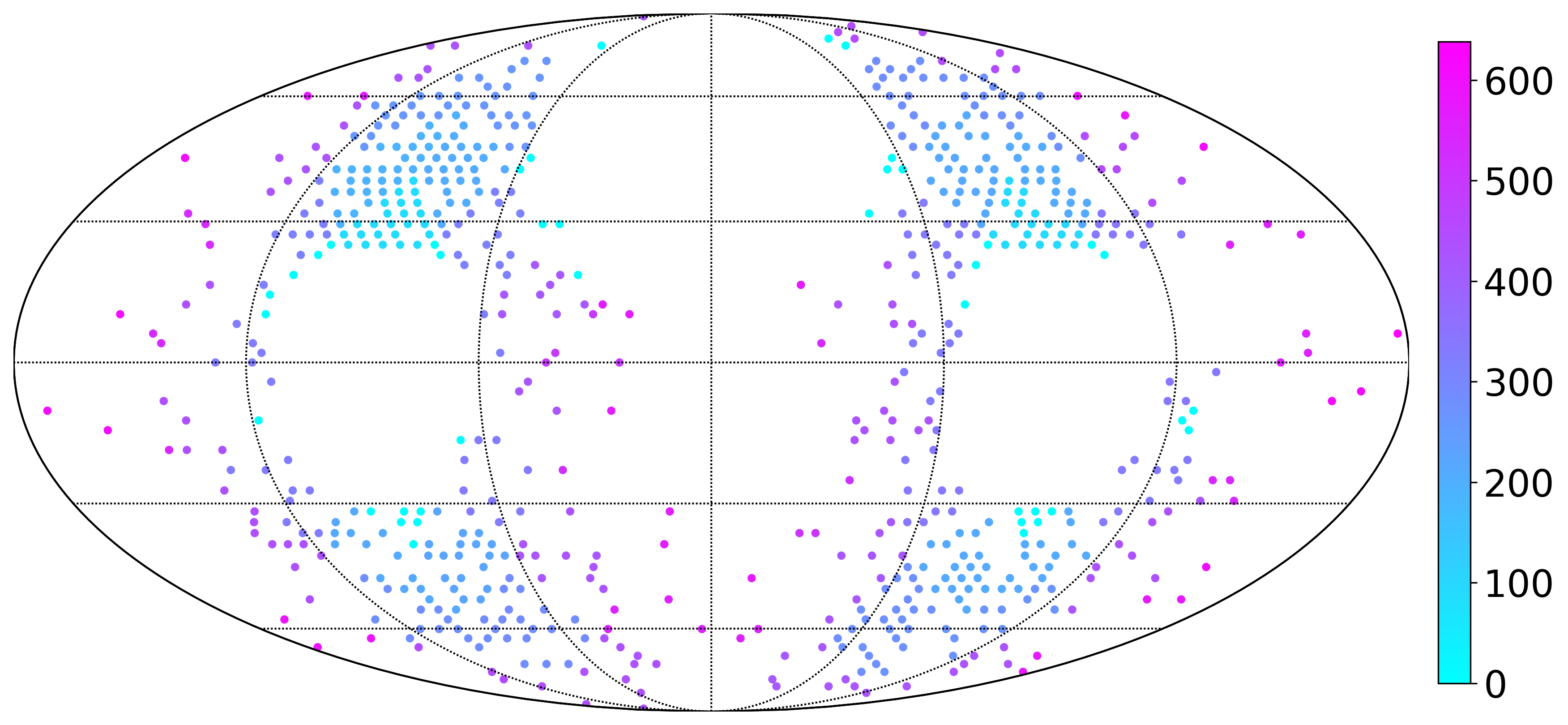}
    \label{fig:hyperparameters-r15-s10-2023}
  }
  \hfil
  \subfloat[$r=15$, $s=20$]{\includegraphics[width=.3\textwidth]{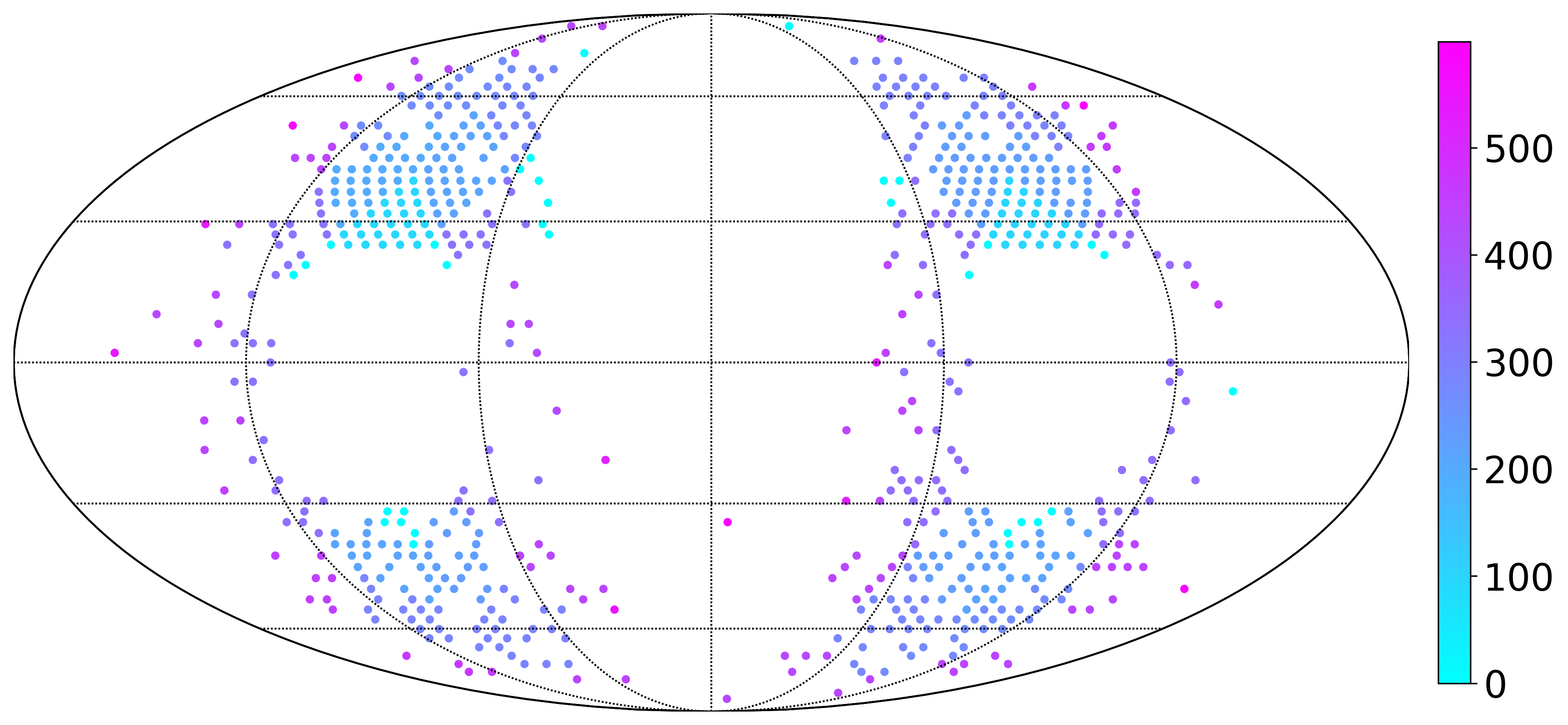}
    \label{fig:hyperparameters-r15-s20-2023}
  }

  \caption{
    \textbf{Trajectory optimization hyperparameters}.
    We executed our trajectory optimization algorithm with different choices for the parameters $r$ and $s$ for sample number 1 and plotted the resulting spherical trajectories.
    In the three rows, we visualized the resulting trajectory when different disk radius parameters ($r=5$, $r=10$ and $r=15$ radians) are applied for the neighborhood of a given pose.
    Poses within this neighborhood were updated with the new score if they were not yet attempted by the robotic arm.
    In the three columns, we visualized the resulting trajectory when different penalty parameters ($s=5$, $s=10$ and $s=20$) are applied on the absorption score.
    $s$ penalizes the absorption rate of the sample from the given angle by decreasing the weight of the given pose in the resulting probability distribution.
  }
  \label{fig:hyperparameters-trajectories-2023}
\end{figure*}

\subsubsection{Disk radius $r$}\label{sec:experiments-score-update-disk-radius-r-2023}
Our choices for the disk radius parameters $\{5,10,15\}$ correspond to $8$, $32$, and $78$ poses on the sphere with high sampling density.
These numbers cover $0.18$\%, $0.74$\%, and $1.8$\% of the sphere surface.
However, more importantly, these numbers cover $19.5$\%, $78$\%, and $190$\% of a single spot on the lower sampled sphere from the scout scan step.
For example, by choosing $r=15$, we would overwrite the resulting scores of two poses from the scout scan with a single measurement in one iteration of the optimization loop.

\subsubsection{Score penalty $s$}\label{sec:experiments-score-update-score-penalty-s-2023}
The score penalty $s$ penalizes the absorption rate of the sample from the given angle by decreasing the weight of the given pose in the resulting probability distribution.
For example, our choices for $s=\{5,10,20\}$ decrease the weight of a pose with score $558$ (higher absorption) from \SI{2.11e-1} ($s=1$) to \SI{4.23e-4} ($s=5$), \SI{1.79e-7} ($s=10$), and \SI{3.20e-14} ($s=20$) compared to a pose with score $118$ (lower absorption).
Considering that $p(i) = \frac{w_{i}}{S}$ where $w_{i}$ is the weight and $S$ is the sum of all weights, we can see that increasing the penalty parameter $s$ yields a significant reduction in the pose's sampling probability.
The scores $118$ and $558$ are the lowest and highest absorption score of the scout scan for sample Nr. 1.
We aim to study the effect of choosing different penalty parameters $s$ on the resulting pose distribution and find a suitable range of choices for $s$.

In Fig. \ref{fig:hyperparameters-trajectories-2023}, we plotted the results of our experiments for the parameters $r$ and $s$.

\subsection{Simulated CT measurements}\label{sec:experiments-measurements-2023}
We conducted six experiments in total: two samples were simulated on three trajectories: whole sphere, random, and optimized.

\begin{figure*}[t]
  \centering
  \subfloat[Spherical mapping, low density, specific pose highlighted]{\includegraphics[width=.33\textwidth]{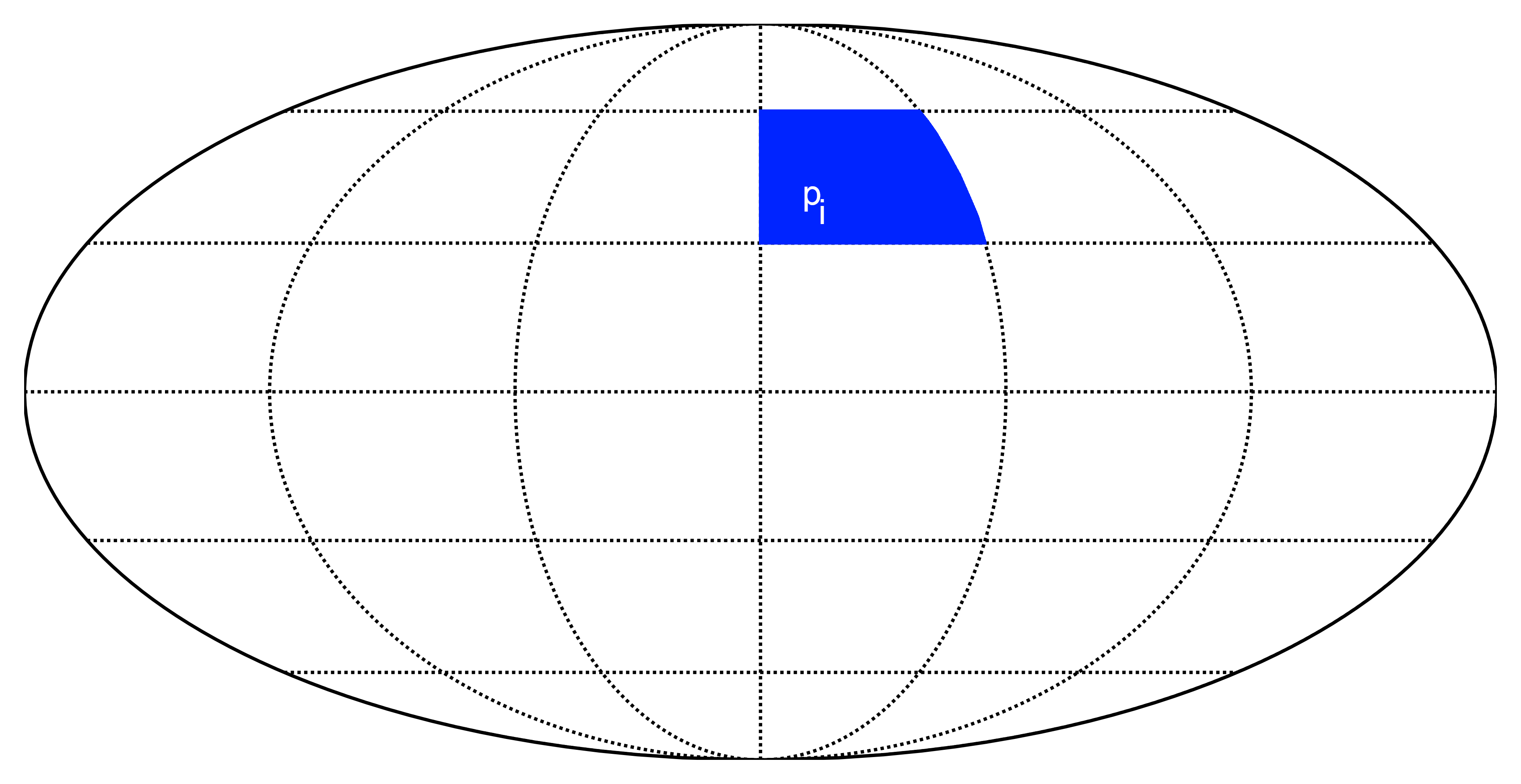}%
    \label{fig:spherical-mapping-low-density-marker-spot-2023}
  }
  \hfil
  \subfloat[Spherical mapping, high density, corresponding poses highlighted]{\includegraphics[width=.33\textwidth]{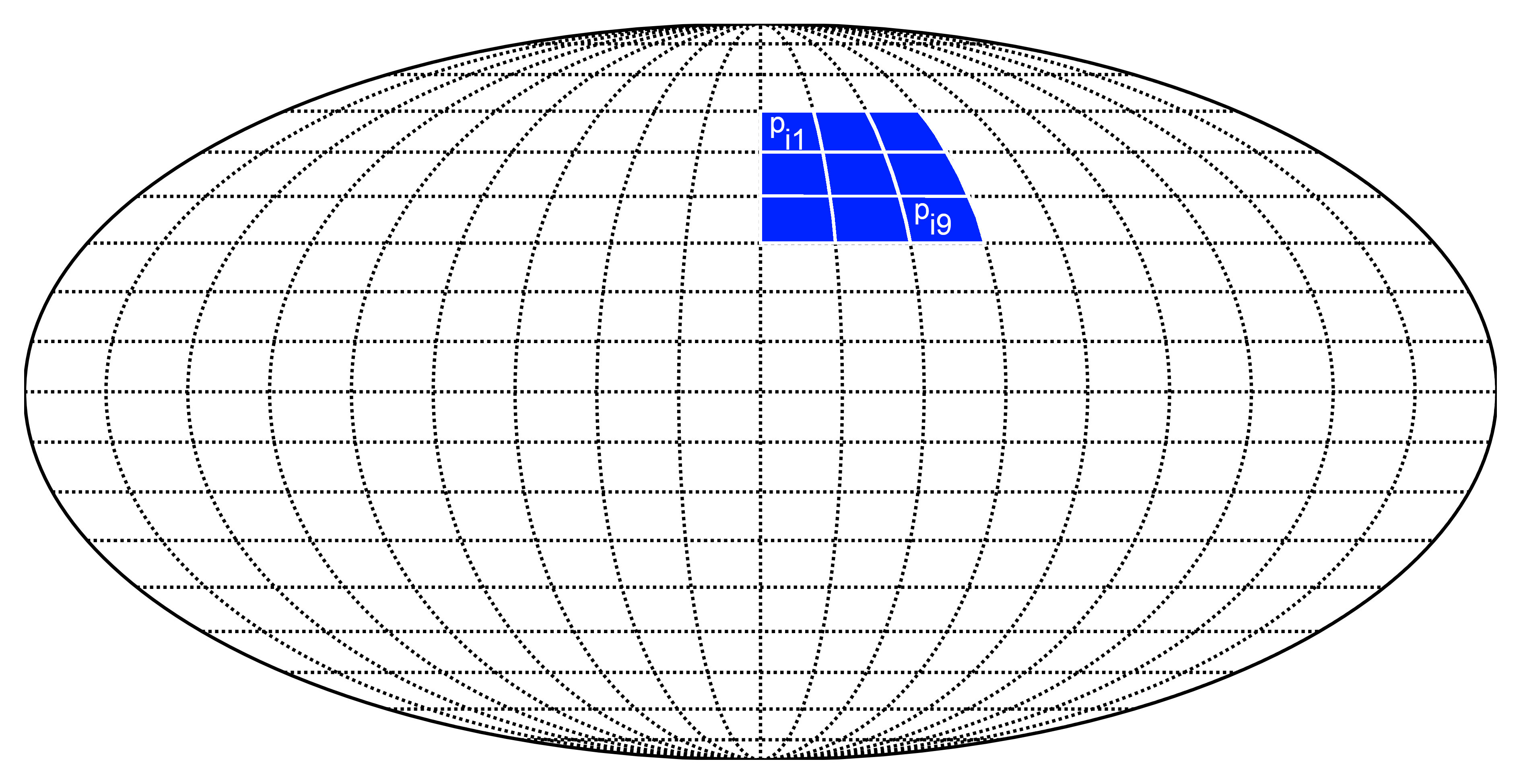}%
    \label{fig:spherical-mapping-high-density-marker-spot-2023}
  }

  \subfloat[Spherical trajectory, poses covering whole sphere]{\includegraphics[width=.3\textwidth]{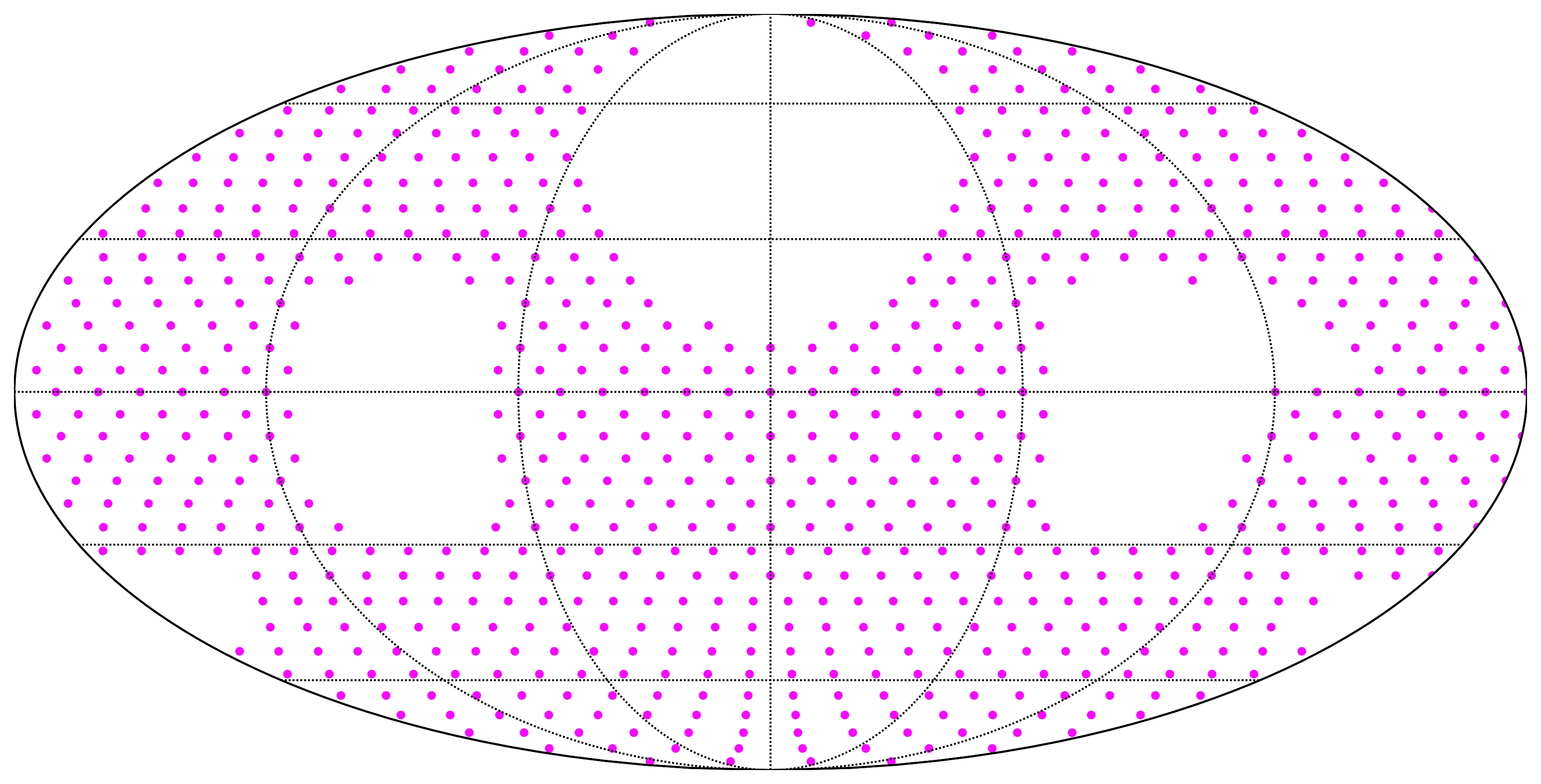}%
    \label{fig:trajectory-sphere-whole-2023}
  }
  \subfloat[Spherical trajectory, random poses]{\includegraphics[width=.3\textwidth]{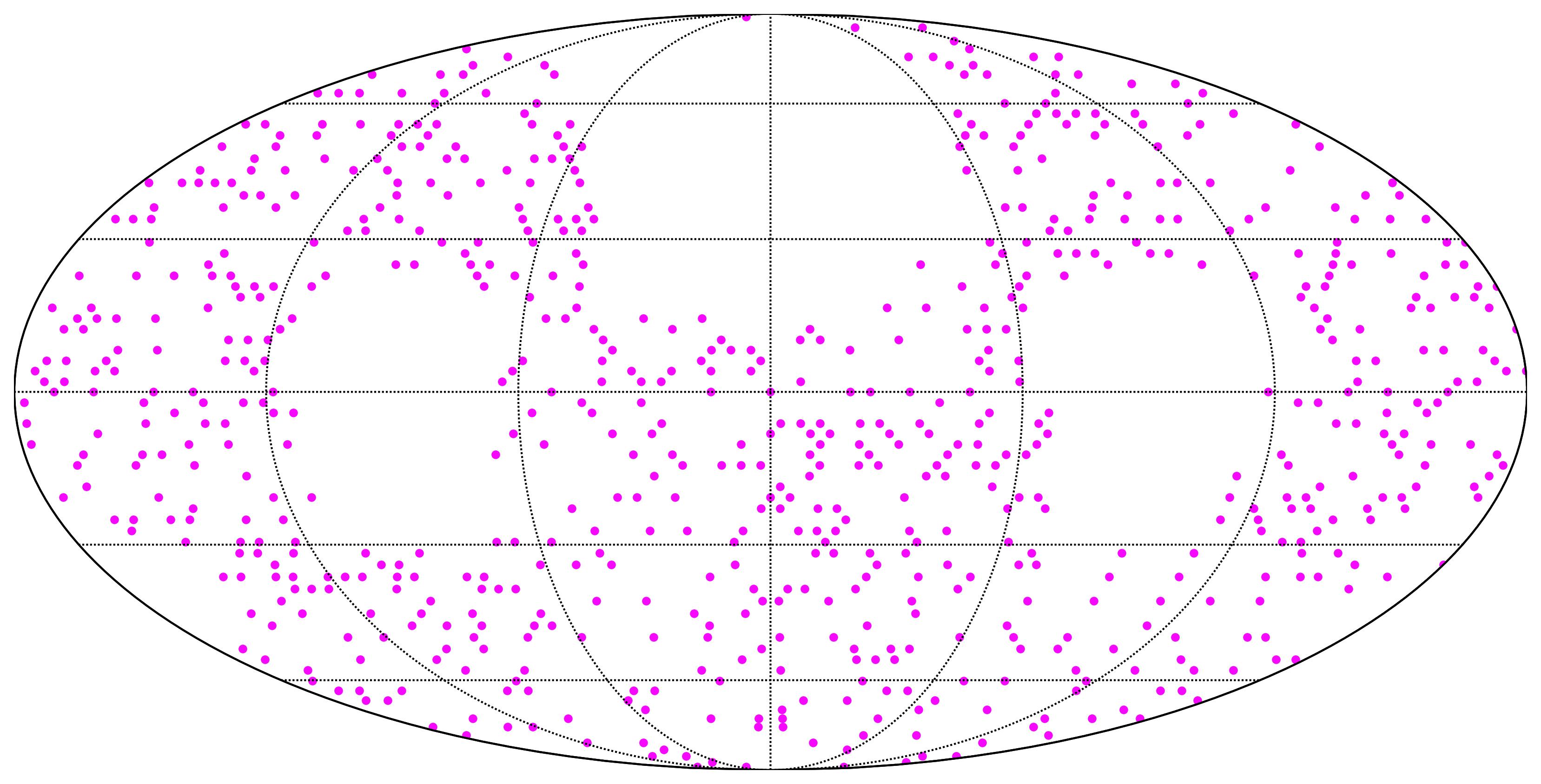}%
    \label{fig:trajectory-sphere-random-2023}
  }
  \subfloat[Spherical trajectory, optimial poses, \\$r=5$, $s=20$]{\includegraphics[width=.3\textwidth]{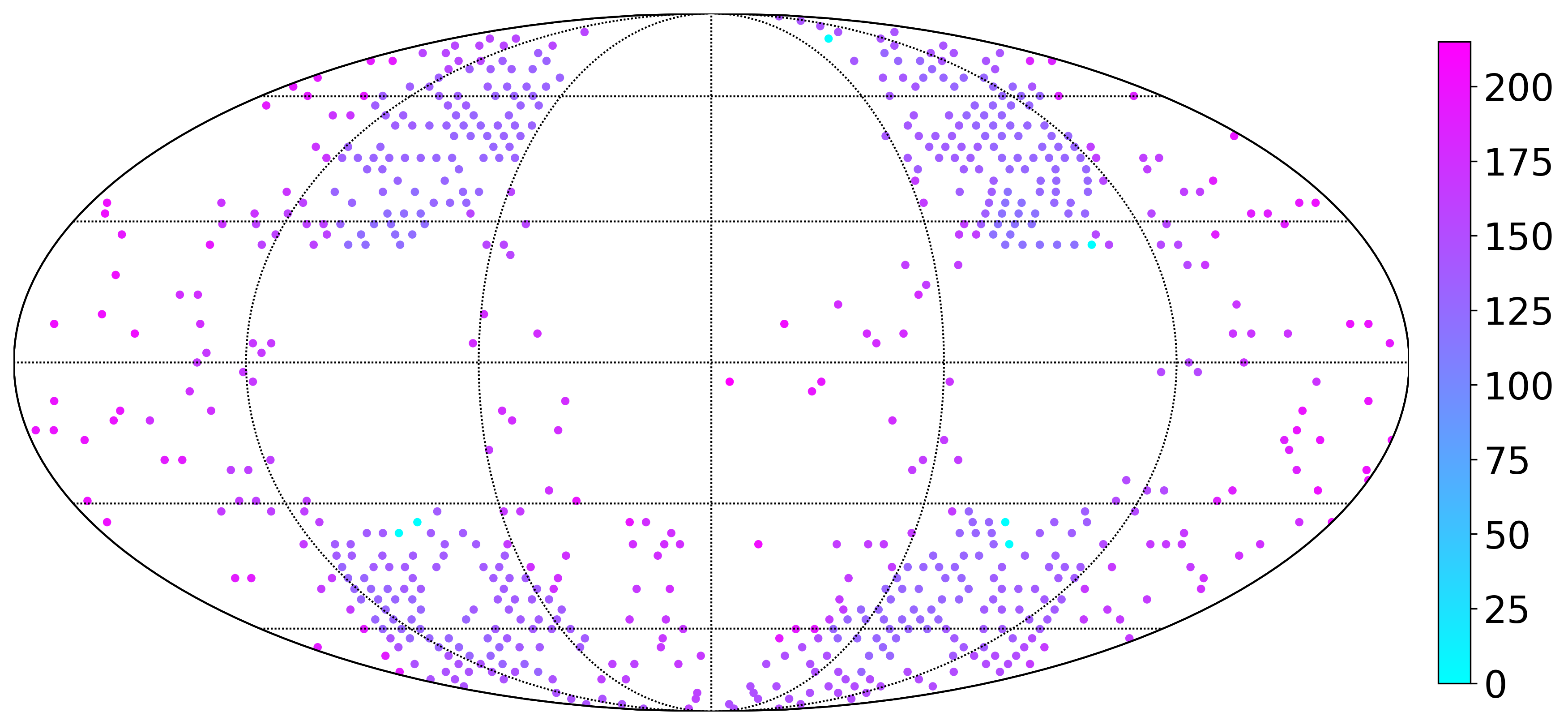}%
    \label{fig:trajectory-sphere-optimized-2023}
  }

  \caption{
    \textbf{Spherical trajectories}.
    In (\subref{fig:spherical-mapping-low-density-marker-spot-2023}) and (\subref{fig:spherical-mapping-high-density-marker-spot-2023}), we visualized the mapping of specific regions between sphere surfaces with low and high sampling density.
    In (\subref{fig:trajectory-sphere-whole-2023}), (\subref{fig:trajectory-sphere-random-2023}) and (\subref{fig:trajectory-sphere-optimized-2023}), we plotted the spherical trajectories for the three different strategies of choosing poses on the sphere for sample no. 2:
    Whole sphere coverage (\subref{fig:trajectory-sphere-whole-2023}), random sampling (\subref{fig:trajectory-sphere-random-2023}) and online optimization based on absorption score of previous measurements (\subref{fig:trajectory-sphere-optimized-2023}).
    In (\subref{fig:trajectory-sphere-whole-2023}), (\subref{fig:trajectory-sphere-random-2023}) the spots on the sphere are colored uniformly as no score was calculated in these cases.
    The resulting image reconstructions for the three trajectories are displayed in Fig. \ref{fig:oto-reconstructions-sample-2-2023}.
  }
  \label{fig:spherical-trajectories-2023}
\end{figure*}

The three trajectories are visualized in figures \ref{fig:spherical-trajectories-2023} (\subref{fig:trajectory-sphere-whole-2023}), (\subref{fig:trajectory-sphere-random-2023}) and (\subref{fig:trajectory-sphere-optimized-2023}).
We generated the first trajectory by uniformly sampling poses on the sphere surface with the HEALPix library with the parameter $N_{side}=9$, which results in $972$ poses.
The resulting trajectory contained fewer poses ($725$) than $972$ as not all regions of the sphere are reachable by the robotic arm (see Fig. \ref{fig:spherical-trajectories-2023} (\subref{fig:trajectory-sphere-whole-2023})).
We generated the second trajectory by first sampling poses on the sphere surface uniformly with a higher density ($N_{side}=19$) and subsequently sampling poses without replacement from the uniformly distributed poses.
The resulting trajectory contained $748$ poses (see Fig. \ref{fig:spherical-trajectories-2023} (\subref{fig:trajectory-sphere-random-2023})).
We were not able to match the number of poses in Fig. \ref{fig:spherical-trajectories-2023} (\subref{fig:trajectory-sphere-whole-2023}) precisely as the motion planning step is executed after sampling the poses, and not all poses are reachable as mentioned earlier.
The third trajectory is generated with the pipeline explained in \ref{sec:pipeline-overview-2023}.
The resulting trajectory contains $760$ poses (see Fig. \ref{fig:spherical-trajectories-2023} (\subref{fig:trajectory-sphere-optimized-2023})).

\begin{figure*}[t]
  \centering
  \subfloat[Sample No. 1]{\includegraphics[width=0.35\textwidth]{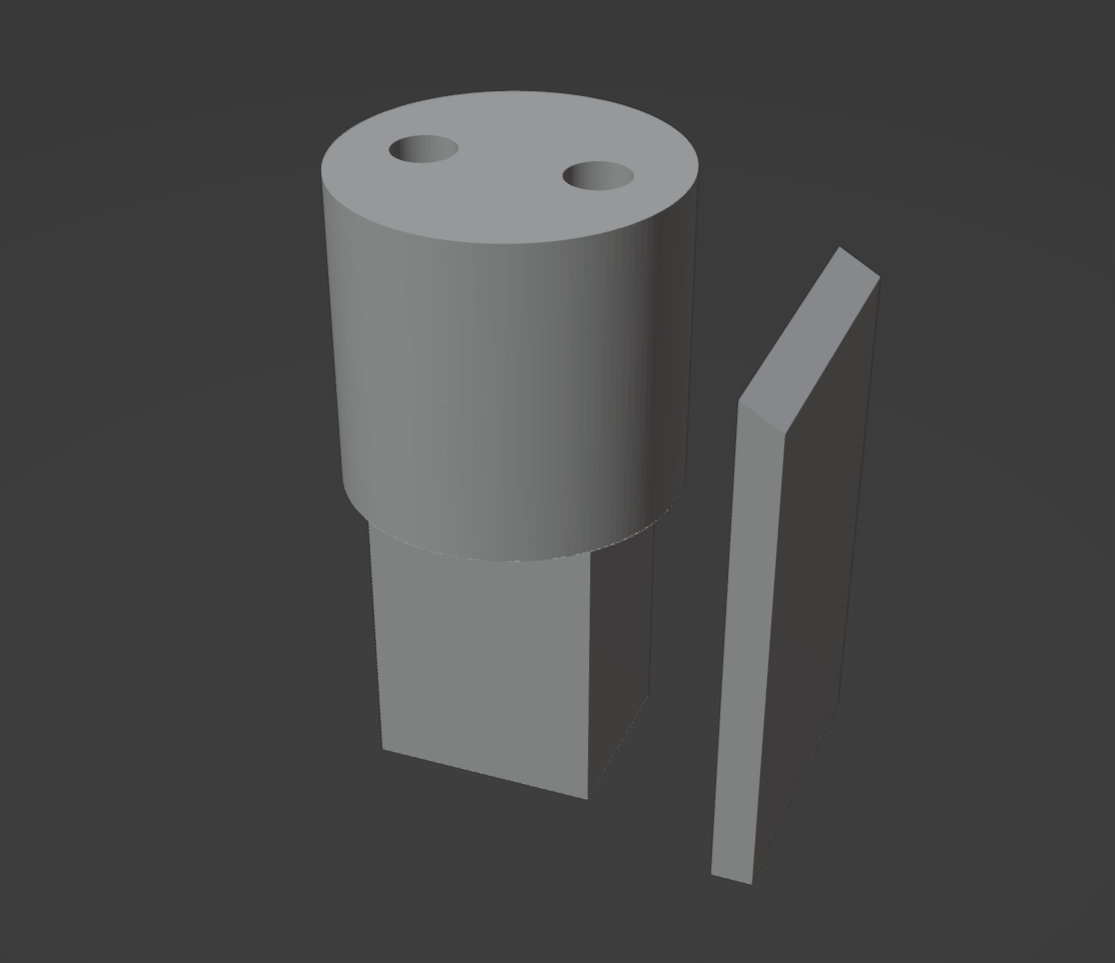}%
    \label{fig:sample-1-2023}
  }
  \hfil
  \subfloat[Sample No. 2]{\includegraphics[width=.35\textwidth]{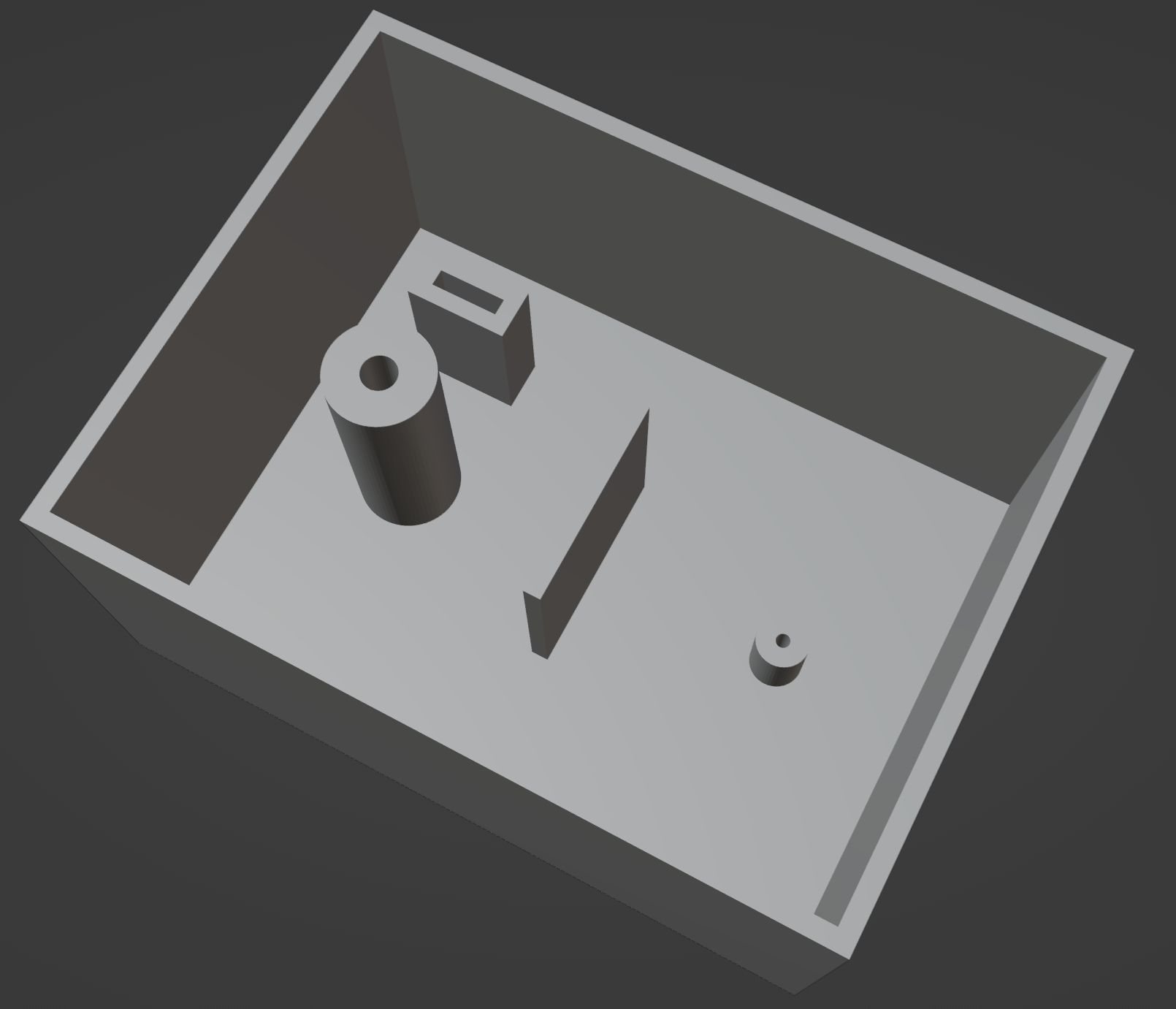}%
    \label{fig:sample-2-2023}
  }

  \caption{
    \textbf{Samples}.
    We modeled two samples, each containing an absorber plate that simulates highly absorbing parts.
    In (\subref{fig:sample-1-2023}), the sample is a combination of a cubic and a cylindrical object stacked vertically (left), placed next to an absorber plate with dimensions 4 x 30 x 50 mm.
    In (\subref{fig:sample-2-2023}), the sample is made up of an open box containing one cubic and two cylindrical objects separated by a single absorber plate with dimensions 1 x 12 x 12 mm.
    For both samples, we modeled holes into the cubic and cylindrical objects for evaluating reconstruction image quality.
  }
  \label{fig:sample-2023}
\end{figure*}

The two samples we used for all experiments are displayed in Fig. \ref{fig:sample-2023}.
Both samples consist of objects with an internal structure and an absorber plate.
We expect the absorber plate to affect reconstruction image quality badly in our measurements due to its higher attenuation coefficient than the sample objects.

The first sample is displayed in Fig. \ref{fig:sample-2023}(\subref{fig:sample-1-2023}).
It combines a cubic and a cylindrical object stacked vertically and placed next to an absorber plate.
The cubic object in the bottom measures 22 x 18 x 30 mm and contains a cylindrical hole with a 7 mm diameter.
The cylindrical object has a diameter of 32 mm with two cylindrical holes, with 6 mm diameter each and a height of 30 mm.
Both objects total 60 mm in height.
We placed the absorber plate with dimensions 4 x 30 x 50 mm next to the sample described above.

The second sample is displayed in Fig. \ref{fig:sample-2023}(\subref{fig:sample-2-2023}).
We modeled this sample as an open box that contains three sample objects, two cylinders with varying heights, and a cubic object.
We placed the absorber plate with dimensions 1 x 12 x 12 mm in the center of the box.
The cubic object at the top left measures 6 x 3 x 10 mm and contains a cubic hole with 1 mm padding to both sides.
The cylindrical object below it measures 6 x 6 x 15 mm and contains a cylindrical hole with a 2 mm diameter.
The cylindrical object to the right of the absorber measures 3 x 3 x 5 mm, and it contains a cylindrical hole with a 0.8 mm diameter.

The two samples include absorbers with different dimensions and thicknesses.
The absorber in sample number 1 has a much greater volume of 6000 mm\textsuperscript{3} compared to the absorber in sample number 2 with just 144 mm\textsuperscript{3}.

We set the absorption coefficient of the absorber plates in the sample objects to $7.0$ and the rest of the samples to $1.5$ for realistic modeling of the ratio between highly absorbing and less absorbing parts of a sample in a laboratory X-ray CT measurement.
For reference: at an X-ray energy of $45$ keV and material thickness of $3$ mm, \textit{aluminium} attenuates $30.63$\% of the X-rays compared to \textit{polyethylene} with $5.87$\% attenuation \cite{helmholtz-calc}.
These values give us a ratio of $5.21$ and hence are comparable to our ratio of $4.67$.

\begin{figure*}[t]
  \centering
  \includegraphics[width=0.8\textwidth]{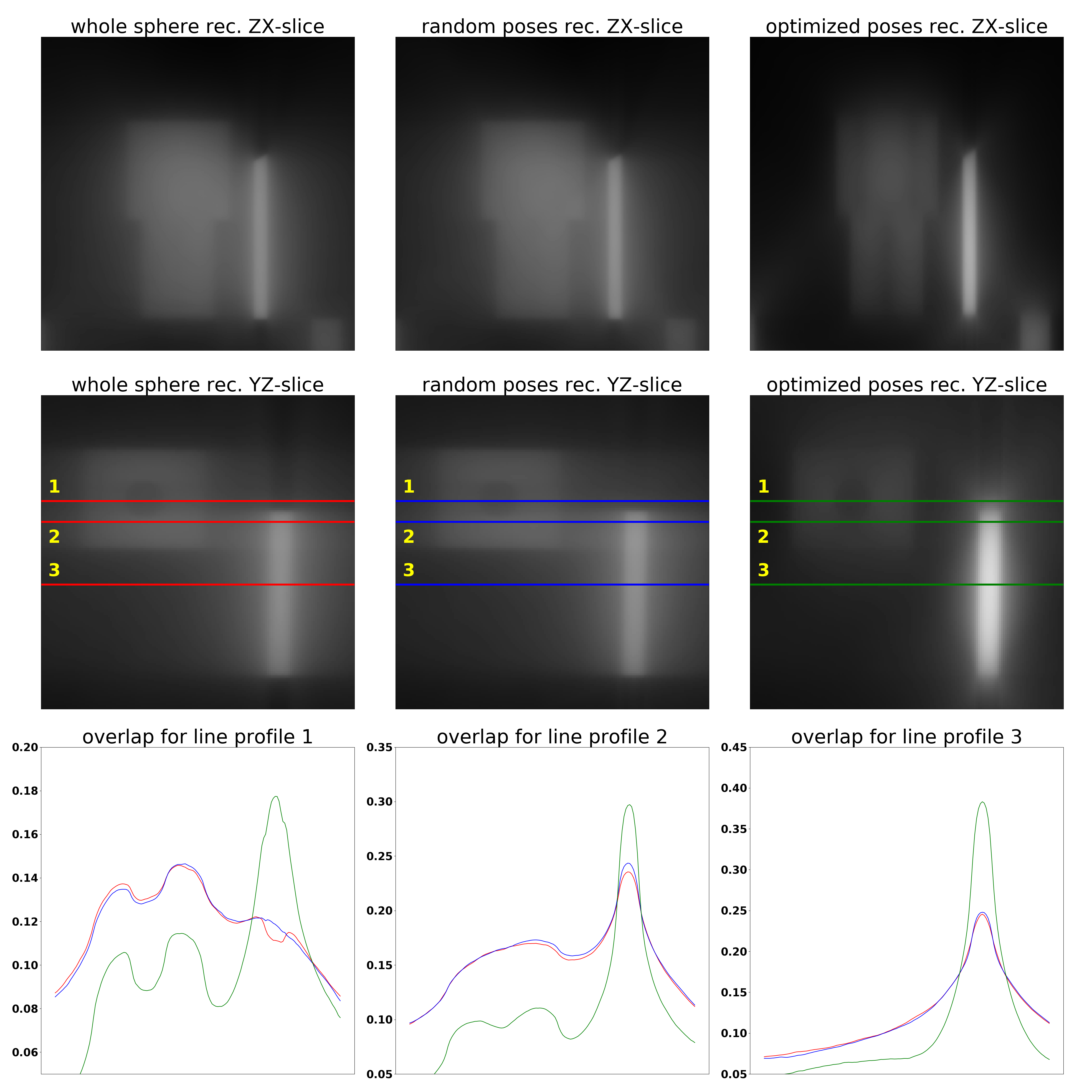}

  \caption{
  \textbf{Experimental results. Sample No. 1}.
  The first sample (see Fig. \ref{fig:sample-1-2023}) was measured on three different types of trajectories (similar to Fig. \ref{fig:spherical-trajectories-2023} ((\subref{fig:trajectory-sphere-whole-2023}), (\subref{fig:trajectory-sphere-random-2023}) and (\subref{fig:trajectory-sphere-optimized-2023}))) and reconstructed with the robotic arm in the simulation environment of our robotic software package \cite{Pekel_2023}.
  The optimized trajectory was generated with the hyperparameters $r=5$ and $s=20$.
  The reconstruction volumes are registered and aligned with our calibration algorithm \cite{Pekel_2022}.
  We binned the detector images with $4*4$, and the reconstruction volume has dimensions $720^{3}$.
  A zoom factor of 4.8x was applied to the slices to crop the region of interest.
  We plotted line profiles at three different positions for the YZ slices.
  }
  \label{fig:oto-reconstructions-sample-1-2023}
\end{figure*}

\begin{figure*}[t]
  \centering
  \includegraphics[width=0.8\textwidth]{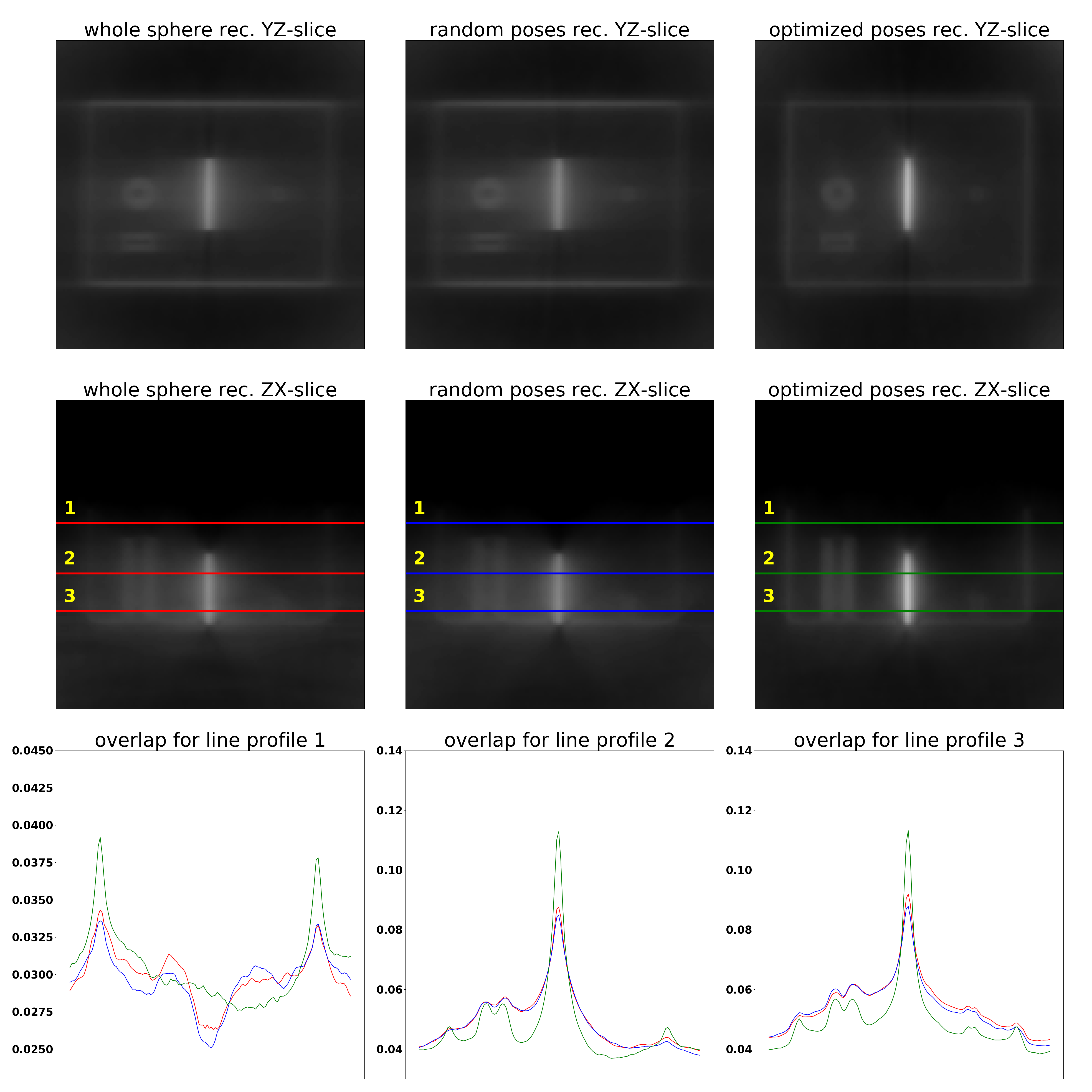}

  \caption{
  \textbf{Experimental results. Sample No. 2}.
  The second sample (see Fig. \ref{fig:sample-2-2023}) was measured on the three trajectories displayed in Fig. \ref{fig:spherical-trajectories-2023} ((\subref{fig:trajectory-sphere-whole-2023}), (\subref{fig:trajectory-sphere-random-2023}) and (\subref{fig:trajectory-sphere-optimized-2023})) and reconstructed with the robotic arm in the simulation environment of our robotic software package \cite{Pekel_2023}.
  The optimized trajectory was generated with the hyperparameters $r=5$ and $s=20$.
  The reconstruction volumes are registered and aligned with our calibration algorithm \cite{Pekel_2022}.
  We binned the detector images with $4*4$, and the reconstruction volume has dimensions $720^{3}$.
  A zoom factor of 4.8x was applied to the slices to crop the region of interest.
  We plotted line profiles at three different positions for the ZX slices.
  }
  \label{fig:oto-reconstructions-sample-2-2023}
\end{figure*}

For the quantitative analysis of our experiments, we have calculated a metric on the reconstructions of our experiments from section \ref{sec:experiments-and-results-2023}.
The trajectories for these experiments are displayed in Fig. \ref{fig:spherical-trajectories-2023} and the qualitative results are illustrated in Figures \ref{fig:oto-reconstructions-sample-1-2023} and \ref{fig:oto-reconstructions-sample-2-2023}.
In Tables \ref{tab:quantitative-statistics-sample-1-2023} and \ref{tab:quantitative-statistics-sample-2-2023}, we have provided statistics on the quantitative difference in the reconstruction images between the three trajectories.
We used the gradient magnitude to measure the sharpness of the line profiles on the resulting reconstruction slices (see Figures \ref{fig:oto-reconstructions-sample-2-2023} and \ref{fig:oto-reconstructions-sample-1-2023}).

\begin{table}
  \centering
  \caption{\textbf{Reconstruction quality statistics}.
    Quantitative analysis of reconstruction image quality for two samples.
    The trajectory optimization parameters $r=5$ and $s=20$ were used for the underlying experiments.
    We calculated the metrics on the line profiles of the YZ (sample no. 1) and ZX-slices (sample no. 2) of the reconstructions (see figures \ref{fig:oto-reconstructions-sample-1-2023} and \ref{fig:oto-reconstructions-sample-2-2023}).
  }

  \begin{tabular}{c | c c c}
    \hline
              & whole sphere & random   & optimized            \\
    \hline
    profile 1 & 0.000991     & 0.000945 & 0.002284 (+130.5 \%) \\
    profile 2 & 0.001965     & 0.002047 & 0.003631 (+84.8 \%)  \\
    profile 3 & 0.002054     & 0.002099 & 0.004346 (+111.6 \%)
  \end{tabular}
  \subcaption{
    \textbf{Sample No. 1}: Gradient magnitude of line profiles plotted in Fig. \ref{fig:oto-reconstructions-sample-1-2023}.
  }
  \label{tab:quantitative-statistics-sample-1-2023}

  \begin{tabular}{c | c c c}
    \hline
              & whole sphere & random   & optimized           \\
    \hline
    profile 1 & 0.000213     & 0.000216 & 0.000283 (+32.9 \%) \\
    profile 2 & 0.000811     & 0.000760 & 0.001466 (+80.8 \%) \\
    profile 3 & 0.000799     & 0.000766 & 0.001356 (+69.7 \%)
  \end{tabular}
  \subcaption{
    \textbf{Sample No. 2}: Gradient magnitude of line profiles plotted in Fig. \ref{fig:oto-reconstructions-sample-2-2023}.
  }
  \label{tab:quantitative-statistics-sample-2-2023}

\end{table}

\section{Discussion}\label{sec:discussion-2023}
In this section, we will discuss the results of our experiments from section \ref{sec:experiments-and-results-2023}, where we aimed to measure our system's performance for different choices of the hyperparameters $r$ and $s$ and the reconstruction image quality.
We categorized the first set of experiments by the choices for the parameters $r$ and $s$ and fixed the sample to sample no. 1.
We fixed the parameters $r$ and $s$ for the second set of experiments and used two different samples to evaluate our algorithm's performance on two different sample inputs.

\subsection{Trajectory optimization parameters}\label{sec:discussion-trajectory-update-parameters-2023}
We executed experiments with different choices for the parameters $r$ and $s$ to determine the effects on the distribution of poses on the sphere.
We plotted the resulting spherical trajectories in figures \ref{fig:hyperparameters-trajectories-2023} (\subref{fig:hyperparameters-r5-s5-2023}) to (\subref{fig:hyperparameters-r15-s20-2023}).

The parameter $r$ is the disk's radius that contains all poses in the neighborhood that are updated with the newly calculated score in the update step explained in section \ref{sec:score-update-2023}.
We chose $r=\{5,10,15\}$ radians as update radiuses in our experiments.
In section \ref{sec:experiments-score-update-disk-radius-r-2023}, we have mentioned that these parameters will update $0.18$\%, $0.74$\%, and $1.8$\% of the entire sphere surface.
We can see in the rows of Fig. \ref{fig:hyperparameters-trajectories-2023} that our choices for $r$ are visible on the trajectories when fixing $s$ by choosing a column and examining different rows for varying $r$.
However, the difference caused by different $r$ will not impact the reconstruction image as the distribution of points only differs in a small neighborhood when $s$ is fixed.
The re-distribution of points in a small neighborhood on a sphere does not cause significantly different angles with higher absorption.

We apply the parameter $s$ as a penalty on the absorption score of a given pose to decrease its weight in the probability distribution.
We chose $s=\{5,10,20\}$ as penalty parameters in our experiments.
We can see in the columns of Fig. \ref{fig:hyperparameters-trajectories-2023} that with increasing $s$, the poses on the spherical trajectory concentrate in four regions of the sphere surface.
The remaining regions are sparsely populated, and poses in these regions are less likely to be sampled with increasing $s$.
The concentration of sampled poses in particular highly absorbing regions of the sphere surface, matches our initial goal of avoiding images with high absorption.
If the parameter $s$ is chosen too high, information about certain angles of the sample will be incomplete or missing for the image reconstruction step.
The information loss will lead to worse image reconstruction quality.

\subsection{Simulated CT measurements}\label{sec:discussion-measurements-2023}
We fixed the trajectory optimization parameters for the CT measurements to $r=5$ and $s=20$.
These parameters improve image reconstruction quality as discussed in section \ref{sec:discussion-trajectory-update-parameters-2023}.
We used two samples (see Fig. \ref{fig:sample-2023}) to evaluate the reconstruction images for image quality and sharpness.

\subsubsection{Image quality}
We can discuss our system's performance from a qualitative and quantitative perspective by examining the results in Figures \ref{fig:oto-reconstructions-sample-1-2023} and \ref{fig:oto-reconstructions-sample-2-2023} and Tables \ref{tab:quantitative-statistics-sample-1-2023} and \ref{tab:quantitative-statistics-sample-2-2023}.

We can divide our observations on the reconstruction image quality into two parts: the region around the absorber and the rest of the sample.
We expect as few artifacts as possible for the region around the absorber.
We expect the internal structures to be clearly visible for the remaining parts of the sample and the outer edges to be sharp.

First, we will analyze the region around the absorber.
For sample no. 1, we can see in Fig. \ref{fig:oto-reconstructions-sample-1-2023} in the first row that the optimized trajectory can avoid most of the artifacts that would otherwise be present in the ZX-slices of the reconstruction.
Compared to the first row, the artifacts are not avoided as well in the YZ-slice (second row) with the optimized trajectory. However, there is still a visible difference compared to the other two trajectories.
With the whole sphere (left column) and random (center column) trajectory types, the region around the absorber has artifacts in all directions.
For sample no. 2, the region around the absorber has significantly fewer artifacts for the optimized trajectory (right column) than the whole sphere and random trajectory, which have artifacts in all directions.
In contrast to sample no. 1, the difference in quality is significant in both slices instead of just one slice.

We will examine the outer edges and internal structures of the cylindrical and cubic objects for the remaining parts of the samples.
For sample no. 1, the optimized trajectory can partly visualize the internal structure of the stacked objects next to the absorber.
The other two trajectory types (left and middle column) can not visualize the internal structure in the stacked object; the object's interior is uniformly white.

For samples no. 1 and 2, we can not observe a difference in quality between the whole sphere and random trajectory for the region around the absorber and the remaining sample parts.

\subsubsection{Image sharpness}
We will evaluate reconstruction image sharpness for samples no. 1 and 2 by interpreting the line profiles plotted in Figures \ref{fig:oto-reconstructions-sample-1-2023} and \ref{fig:oto-reconstructions-sample-2-2023} and the corresponding statistics in Tables \ref{tab:quantitative-statistics-sample-1-2023} and \ref{tab:quantitative-statistics-sample-2-2023}.

In general, when looking at the line profiles for both samples, the apparent spikes are steeper and have a higher amplitude with the optimized trajectory (green lines) than the profiles of the other two trajectories (red and blue lines).
There is an exception to this statement when the line profiles pass the area above the absorber on the YZ-slices for sample no. 1.
Here we can see slight artifacts for the optimized trajectory when there are no visible artifacts for the other two trajectories.
We suspect that acquisition angles parallel to the absorber are causing these, which is expected because these angles have a minor absorption score with the L0-norm.
The minor score comes from the fact that the projected area of the absorber is smaller on the sinogram compared to an angle that captures the absorber head-on.
This behavior can be fine-tuned by decreasing the penalty parameter $s$.

Another critical observation on the line profiles is the appearance of the samples' smaller inner structures in the optimized trajectory profiles.
The profiles of the whole sphere and random trajectories do not correctly capture these regions.
For example, on the overlapping plot for line profile 2 of sample no. 2 (green line), we can see the correct representation of the cylindrical hole inside the cylindrical part to the left of the absorber and the wall structure next to the sample.
The whole sphere and random trajectories (blue and red lines) cannot represent this information accurately.

We can see in Tables \ref{tab:quantitative-statistics-sample-1-2023} and \ref{tab:quantitative-statistics-sample-2-2023} that with the optimized trajectory, our gradient magnitude metric is improved by up to $130.5$\% and $80.8$\% on the image reconstructions for sample no. 1 and 2.

In our previous publication, we used the same metric for comparing the difference in sharpness between the image reconstructions of measurements on a circular and spherical trajectory in a laboratory X-ray environment \cite{Pekel_2023}.
There, the gradient magnitude improved by $50$\% to $80$\% with a spherical trajectory that covers the whole sphere as illustrated in Fig. \ref{fig:trajectory-sphere-whole-2023}.
This improvement was since the circular trajectory only covered a fraction of the sphere surface compared to the newly introduced spherical trajectory.
With the algorithm introduced in this paper, we are optimizing the spherical trajectory from \cite{Pekel_2023}.
We can conclude from the numbers in Tables \ref{tab:quantitative-statistics-sample-1-2023} and \ref{tab:quantitative-statistics-sample-2-2023} that with an optimized trajectory, we can improve the reconstruction image sharpness even further (up to $130.5$\% with sample no. 1, profile 1).

\subsection{Future work}\label{sec:discussion-future-work-2023}
In future work, we plan to improve the proposed trajectory optimization algorithm in two ways.

First, we plan to parallelize the existing implementation so that the score update (12) and pose sampling (7) steps inside step \textbf{III} of our optimization pipeline in Fig. \ref{fig:oto-pipeline-2023} are executed based on the image reconstruction and segmentation of the previous iteration.
We aim to save execution time for the reconstruction and segmentation step and, more importantly, increase the resolution of the intermediate reconstruction (10) in step \textbf{III}.

Second, we plan to adapt the pose sampling procedure (7) in step \textbf{III} in Fig. \ref{fig:oto-pipeline-2023} for better coverage of areas that were marked as not reachable when the trajectory for the scout scan was generated in \textbf{I} (1).
The current strategy omits comparably large regions on the sphere surface because an unreachable pose of the sphere with low density and hence large regions from step \textbf{II} covers them.
With the new sampling approach, we can explore parts of these regions (e.g., borders of the unreachable poses) and increase image quality by potentially acquiring additional favorable angles in the optimization loop.

The experiments and results in this work are based on simulated studies but should transfer to experimental measurements in a laboratory environment without restrictions.
To verify the applicability of our methods, we plan to execute the experiments from section \ref{sec:experiments-and-results-2023} in our laboratory environment in the future.

\section{Conclusion}
In this work, we have demonstrated using a seven DoF robot as a sample holder for acquiring optimized X-ray computed tomography trajectories using our unified software package with path planning and calibration.
In the introductory paragraphs, we stated that our trajectory optimization algorithm would improve the image quality of 3d reconstructions for complex samples that contain highly absorbing parts by generating optimized and sample-specific trajectories.
We tested our algorithm with samples of different sizes, shapes, and absorption characters, and our findings have confirmed that our algorithm can generate trajectories where the image quality is superior qualitatively and quantitatively compared to a conventional spherical trajectory.

\bibliography{references.bib}{}
\bibliographystyle{IEEEtran}

\end{document}